\documentclass{MITcsail}

\title{Toward a foundation model for forest point clouds}

\author
{Yuanwen~Yue$^{1, 2}$,  
Stefano~Puliti$^{3}$\footnotemark[2],
Damien~Robert$^{4}$\footnotemark[2],
Atakan~Topaloğlu$^{1}$,
Binbin~Xiang$^{3}$, 
Maciej~Wielgosz$^{3}$, 
Jan~Dirk~Wegner$^{4}$,
Rasmus~Astrup$^{3}$, 
Christian~Rupprecht$^{2}$, 
Konrad~Schindler$^{1}$
\\
\vspace{1em} %
\normalfont{\small $^{1}$ETH Zurich}\\
\normalfont{\small $^{2}$University of Oxford}\\
\normalfont{\small $^{3}$Norwegian Institute of Bioeconomy Research (NIBIO)}\\
\normalfont{\small $^{4}$University of Zurich} 
\vspace{2em}
}
\usepackage{amssymb}
\usepackage{amsmath}

\usepackage{xcolor}
\usepackage{colortbl}
\usepackage{booktabs}
\usepackage{float}
\usepackage{multirow}
\usepackage{longtable}

\usepackage{subcaption}
\usepackage{hyperref}
\usepackage{cleveref}
\usepackage{tabularx}

\definecolor{tobecheckedcolor}{rgb}{0.00,0.50,0.60}

\definecolor{rankgreen}{HTML}{78DEAA}

\begin{document}

\footnotetext[2]{These authors contributed equally to this work.}

\maketitle
\thispagestyle{empty} %

\renewcommand{\thefootnote}{\arabic{footnote}}
\setcounter{footnote}{0}

\begin{abstract}
Forest inventories increasingly rely on artificial intelligence (AI) models to derive forest attributes from large-scale 3D point clouds. 
Current models are typically specialized to a single task, sensor, and forest type, making adaptation expensive in terms of annotations, computation, and expertise.
We ask whether a single pretrained model can instead learn transferable representations across diverse forest inventory settings.
Inspired by recent developments in language modelling and computer vision, we take a step toward a \textit{foundation model} (FM) for 3D forestry.
Using LitePT as backbone, we first establish a strong supervised baseline that sets a new state of the art on forest semantic and instance segmentation, tree species classification, and age regression benchmarks.
We then curate a large-scale unlabelled corpus spanning airborne, UAV, and mobile laser scanning across diverse forest ecosystems, and pretrain the same backbone using self-supervised learning.
We systematically evaluate representation learning strategies  by comparing training from scratch, supervised pretraining, and self-supervised pretraining across four representative forestry tasks, under varying annotation budgets.
Compared with training from scratch, self-supervised pretraining accelerates model convergence and consistently improves performance when annotations are scarce.
Compared with task-specific supervised pretraining, self-supervised pretraining yields more transferable representations across downstream forestry tasks.
These findings identify the practical regime in which pretrained representations are most valuable and suggest that instance discrimination, rather than forest semantics, is the main remaining obstacle to a general-purpose 3D forest foundation model.
Code and models are available at: \href{https://github.com/prs-eth/ForPT}{https://github.com/prs-eth/ForPT}.
\end{abstract}

\section{Introduction}
\label{sec_introduction}

Forests play a central role in the global carbon cycle
~\citep{mo2023integrated,pan2024enduring}, biodiversity conservation~\citep{migliavacca2021three,skidmore2021priority}, climate regulation~\citep{forzieri2022emerging,anderegg2022climate}, and the global bioeconomy through timber production and forest-based industries~\citep{hua2022biodiversity,zhang2026forest}, making accurate and scalable forest monitoring a scientific and societal priority. 
Advances in remote sensing technology have enabled the observation of forests in great detail and at unprecedented scale. Satellite imagery, laser scanning (LiDAR) and synthetic aperture radar (SAR) make it possible to acquire measurements from individual trees to entire ecosystems~\citep{duncanson2022_gedi_agbd,lang2023high,fogel2025open,maeda2025expanding,nadeem2026interdisciplinary}. In particular, 3D point clouds acquired by laser scanning capture detailed geometric properties of forests that can support fine-grained analysis of canopy structure~\citep{liu2022novel,zhang2024robust}, biomass~\citep{oehmcke2024deep,borsah2023lidar}, species composition~\citep{michalowska2021review,yip2024community,puliti2025_forspecies20k}, and semantics~\citep{xiang2024_forainet,wielgosz2024_segmentanytree,xiang2025forestformer3d}.
Such 3D data have become central to modern forest inventory and monitoring. 

Recent advances in forest-specific 3D computer vision are expanding the possibilities for fine-grained characterisation of forest ecosystems. A large part of the associated point cloud analysis can be understood as instances of well-studied machine learning tasks. \emph{Semantic segmentation} assigns each point to a semantically meaningful class (e.g., ground, stem/wood, foliage, understory vegetation, or finer taxonomies), supporting component-wise measurements and ecological interpretation \citep{puliti2023_forinstance_paper,shao2024large,lu2025towards,laino2025_segmentedforests}. \emph{Instance segmentation} separates the observed scene into individual trees, a prerequisite for determining tree-level attributes such as location, height, crown volume or diameter at breast height (DBH) \citep{puliti2023_forinstance_paper,xiang2024_forainet,xiang2025forestformer3d,cherlet2026benchmarking}. \emph{Classification} predicts attributes of single trees, like species or functional type \citep{xiang2024_forainet,puliti2025_forspecies20k}. \emph{Regression} directly estimates tree- or stand-level biophysical variables like age or density \citep{PULITI2026115462}.

The last few years have seen rapid progress in supervised deep learning tailored to forest point clouds, along with benchmark datasets that made it possible to quantitatively compare different methods. \citet{xiang2024_forainet} developed ForAINet, a multi-task deep learning pipeline to segment high-density airborne LiDAR and derive tree/stand attributes from the segmented outputs. SegmentAnyTree by~\citet{wielgosz2024_segmentanytree} targets a long-standing operational gap, transferability across different sensing platforms, by training a sensor-agnostic model. They also systematically analyze the performance for varying point densities and carrier platforms (ULS/TLS/MLS). With TreeLearn, \citet{henrich2024treelearn} focus on instance segmentation from ground-based  point clouds (including MLS) and highlight practical issues due to dense stands and crown overlap, as well as the impact of fine-tuning with a small amount of manual labels. ForestFormer3D~\citep{xiang2025forestformer3d} advances end-to-end learning for joint semantic and instance segmentation, with an emphasis on generalization to unseen forest regions and scan setups. 
On the data side, SegmentedForests~\citep{laino2025_segmentedforests} addresses the scarcity of labelled data for \emph{ground-based} analysis by releasing a large, curated TLS/MLS dataset of semantically annotated forest plots. 
FOR-instance~\citep{puliti2023_forinstance_paper,puliti2023_forinstance_data} provides a curated laser scanning benchmark with manual instance annotations and semantic classes, designed to standardize the evaluation of tree segmentation methods across multiple forest types and regions. 
PureForest~\citep{gaydon2025pureforest} provides a benchmark dataset for tree species classification at the forest patch level using ALS point clouds and aerial images.
FOR-species20K~\citep{puliti2025_forspecies20k} targets tree species classification at scale by collating and standardizing more than $20$k taxonomically diverse individual-tree point clouds. 
FOR-age~\citep{PULITI2026115462} offers a high-density laser-scanning dataset of individual trees with annotated age and explores the application of deep learning methods for tree age prediction.

Despite these advances, current models tend to be developed and fitted from scratch for specific variables of interest and for particular forest characteristics, and struggle to generalize. The paradigm of \emph{foundation models} tries to overcome the narrow specialization and instead provide models that can handle a variety of ecosystems, scanning setups and analysis tasks. 
Broadly speaking, there are two ways to construct a foundational model. The first is to collect and annotate training data from diverse forest types and sensing conditions, such that data-driven learning will naturally lead to a model able to operate across the entire range of conditions encountered during training. This approach has been employed successfully in domains like image understanding, where large annotated data collections are relatively easy to assemble~\citep[e.g.,][]{kirillov2023segment,yang2024depth}. Recent data collection efforts like FOR-instanceV2 and FOR-species20K aim in that direction and could form a starting point to produce foundation models for forest point clouds. Still, the classical fully supervised approach is hampered by the large effort needed to produce high-quality annotations, which is not only labour-intensive but also requires specific expertise. 
To alleviate this bottleneck, a recent line of work has developed synthetic forest point cloud generation pipelines and demonstrated that supervised pretraining on these synthetic datasets can reduce the amount of real labelled data required for forest segmentation tasks ~\citep{she2026scaling,liu2026synthetic,jiang2026cross}. 
Nevertheless, the synthetic-to-real transfer gap remains an open challenge, and it is unclear whether representations learned from synthetic data for segmentation transfer effectively to other downstream forestry tasks. Moreover, for many target variables and geographies, labelled ground-truth data remain scarce, particularly when it comes to more subtle ecological attributes like tree age. In contrast, unlabelled point clouds are nowadays abundant due to sensor automation. The second strategy, which forms the basis of the recent AI revolution in disciplines like language processing~\citep[e.g.,][]{achiam2023gpt,comanici2025gemini,liu2024deepseek} and computer vision~\citep[e.g.,][]{radford2021learning, oquab2023dinov2,simeoni2025dinov3}, relies on self-supervised learning, where models are pre-trained on even larger collections of data \emph{without} labels to learn their inherent patterns and structures. The goal is a model that is able to operate across a broad range of forest and sensing conditions and that abstracts the input point cloud into an \emph{embedding}; i.e., a generic representation that retains its characteristic, informative features, such that only a small amount of annotated data is needed to adapt it to a concrete forest type and analysis task.

We note that a similar trend towards large-scale, self-supervised pretraining has emerged in satellite remote sensing~\citep{brown2025alphaearth,jakubik2025terramind,astruc2025anysat,plekhanova2025ssl4eco,szwarcman2025prithvi,guo2024skysense,tolan2024very,bodnar2025foundation}.
As part of the same trend, image-based methods in forestry have started to adapt pretrained foundation models, for instance for disturbance mapping from Sentinel-1 radar imagery~\citep{tian2025vision} and for tree crown segmentation in ground-based~\citep{duguay2026selvamask} as well as aerial imagery~\citep{teng2025sam}.
Moreover, neural network models for satellite imagery have been pretrained specifically for forestry applications, e.g.,~\citet{tolan2024very} train an encoder for VHR Maxar imagery, followed by a dense prediction decoder that outputs canopy height maps.  FoMo~\citep{bountos2025fomo} follows the masked autoencoding paradigm~\citep{he2022masked} to pretrain a forest foundation model for multiple remote sensing modalities, including optical and multispectral satellite imagery, aerial imagery, and synthetic aperture radar (SAR). SSL4Eco~\citep{plekhanova2025ssl4eco} explores the pretraining of ecological representations from satellite imagery.

Compared with imagery, large-scale representation learning for 3D forest point clouds remains far less explored, despite the central role of LiDAR in practical forest economy and ecology. Notable recent exceptions include \citet{opler2026prototree} and \citet{rizaldy2026label}, which we discuss in more detail below.
In the present work, we curate a large unlabelled dataset of forest point clouds
and investigate both fully supervised and self-supervised training of point cloud encoders  across a range of forest types and mapping tasks.
We base our models on LitePT~\citep{yuelitept2026}, a state-of-the-art architecture for 3D point cloud processing, and refer to the resulting family of forest-pretrained models as ForPT.

Our experiments span four forest-related point cloud understanding tasks, including semantic segmentation, instance segmentation, tree age prediction, and tree species classification. 
For each task we consider three different settings: 
(i) the conventional, fully supervised learning from the currently available, public labelled datasets; 
(ii) the strict foundation model setting, where the model is pretrained with self-supervision, then frozen and used to extract an embedding that is fed to a task-specific head, typically a linear layer, a.k.a.\ ``linear probing''; 
(iii) the combined setting, where the pretrained foundation model is fine-tuned in fully supervised fashion together with the output head.
The experiments indicate that fully supervised training from scratch can be viable if diversity is moderate (e.g., only European temperate forest) and sufficient data is available (e.g., $\approx$10,000 segmented and labelled tree instances); but that representations derived with self-supervised learning transfer effectively and boost performance especially in the low-data regime. 
Furthermore, among pretraining strategies, a single self-supervised pretrained backbone yields stronger cross-task transfer than task-specific supervised pretraining.
We conclude that self-supervised learning is an important ingredient for the next step: to create truly foundational forest point cloud models, applicable at continental or global scale, with realistic annotation budgets.
The robustness of self-supervised representation learning is particularly encouraging with a view towards practical applications, where neither the volume of reference annotations nor the compute resources may be available to train large contemporary AI models from scratch.
The main contributions of this study are fourfold:

\begin{itemize}
\setlength{\itemsep}{1pt}
  \item We construct a large-scale, curated dataset of unlabelled forest LiDAR point clouds for self-supervised pretraining, and establish a dedicated evaluation protocol for 3D forestry representation learning.
  \item Using the LitePT backbone, we establish a new state of the art for forest semantic and instance segmentation, tree species classification, and tree age regression under supervised training.
  \item We pretrain the same backbone on unlabelled 3D forest point clouds using a self-distillation approach and systematically compare it against training from scratch and supervised pretraining under varying annotation budgets.
  \item We show that self-supervised pretraining is most effective in low-data regimes and learns more transferable representations than task-specific supervised pretraining across four representative forestry tasks, informing the design of future 3D forest foundation models.
\end{itemize}
\noindent\textbf{Recent related work.} A concurrent study by \citet{opler2026prototree} investigates self-supervised learning on single-tree point clouds for above-ground biomass (AGB) estimation. More closely related to our work, a concurrent effort by \citet{rizaldy2026label} also pretrains models on unlabelled forest LiDAR and evaluates them under reduced supervision. They reach similar conclusions: pretraining is most beneficial when labels are scarce, while current self-supervised objectives do not yet yield strongly instance-discriminative representations.
Notably, the findings are in agreement despite marked technical differences: \citet{rizaldy2026label} pretrain separate sparse U-Net models with a contrastive learning objective for forest segmentation and tree species classification; whereas we pretrain a point transformer with self-distillation, employ a \emph{single} encoder across all four tasks, and evaluate fully supervised pretraining under the same protocol.%

\section{Data}
\label{sec_data}

\begin{figure}[ht]
\centering
\includegraphics[width=\linewidth]{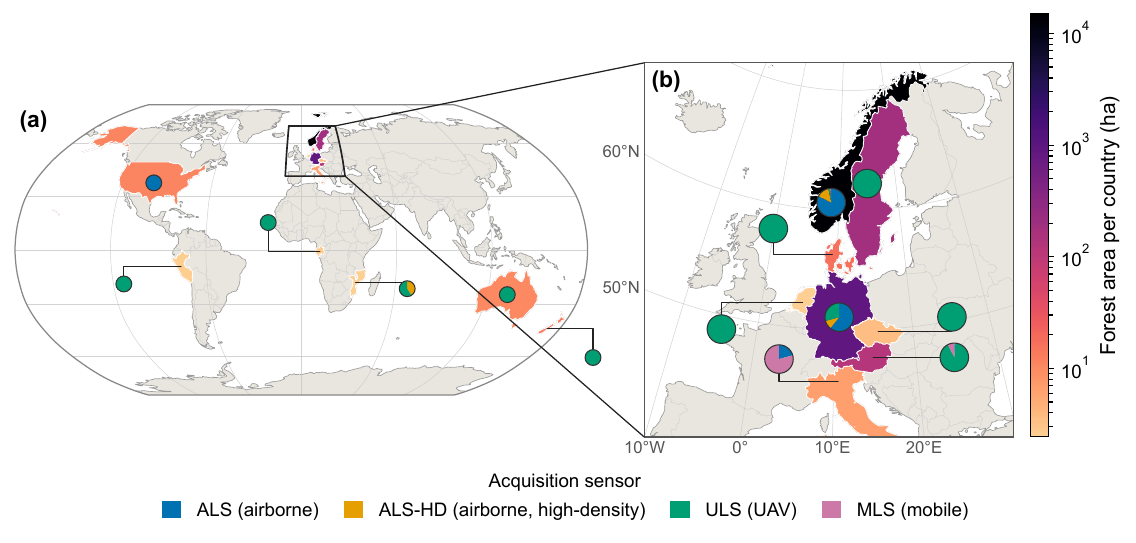}
\caption{\textbf{Geographic distribution of the self-supervised pretraining data.} Each country is coloured by the total forest area it contributes to the pretraining set, and the pie on each country shows how that area breaks down by acquisition sensor. (a) Global view; (b) zoom on Europe.}\label{fig:ssl_chunk_map}
\end{figure}

\subsection{Unlabelled data for self-supervised learning}

\noindent\textbf{Data sources.} To compile a large and diverse collection of forest LiDAR datasets for self-supervised learning, we relied both on NIBIO legacy datasets and on existing open data available from public repositories. The resulting corpus spans multiple geographic regions, forest types, sensor modalities and acquisition conditions. The dataset covers a total forest area of approximately 160 km$^2$ of forest ($\approx15,900$ ha) and contains $\approx33$ billion LiDAR points, with an average point density of $\approx200$ points / m$^2$. The pretraining corpus is disjoint from downstream task data, preventing data leakage between pretraining and downstream evaluation.

\noindent\textbf{Geographical distribution.} The pretraining sites are geographically distributed across multiple continents, including Europe, North America, South America, Africa, and Oceania. Most sites are concentrated in Europe, where dense coverage from different acquisition platforms enables diverse forest conditions and scanning scenarios to be represented. Additional sites outside Europe increase environmental variability. The dataset includes various forest types, including alpine/montane, boreal, temperate, and tropical ecosystems, ensuring diversity in vegetation structure and species composition.

\noindent\textbf{Multi-modality and sensor diversity.} Our collection integrates multiple LiDAR acquisition modalities, including airborne laser scanning (ALS), mobile laser scanning (MLS), and unmanned laser scanning (ULS). This diversity exposes the model to varying point densities, noise characteristics, and viewpoints, facilitating the development of sensor-agnostic representations.
Among these modalities, ALS constitutes the primary data source due to its scalability and ease of acquisition compared with other sensing platforms, making it particularly well suited for large-scale self-supervised pretraining.

\noindent\textbf{Preprocessing.}  Due to variations in sensing platforms, acquisition conditions, some point clouds contain substantial noise and isolated artifacts. To improve data quality and consistency, we apply statistical outlier removal as a preprocessing step. The unlabelled forest dataset consists of large-scale point cloud tiles, each typically covering spatial extents of $200~\mathrm{m} \times 200~\mathrm{m}$, which are computationally prohibitive to process directly with deep learning models. We therefore subdivide each tile into smaller $20~\mathrm{m} \times 20~\mathrm{m}$ spatial patches.

\begin{figure}[!t]
\centering
\includegraphics[width=\linewidth]{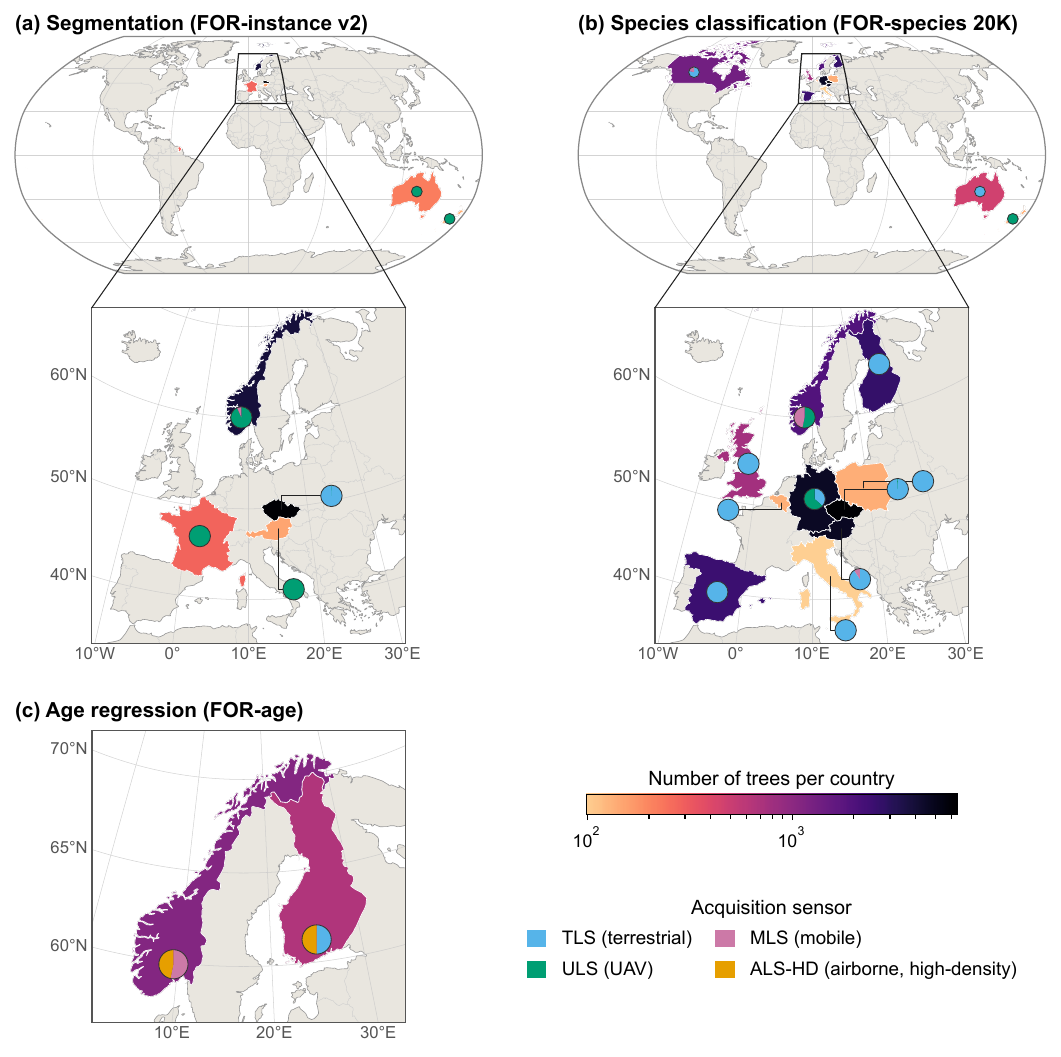}
\caption{\textbf{Geographic distribution of the downstream-task datasets.} Each country is coloured by the number of labeled trees it contributes, and the pie on each country shows the split across acquisition sensors.}\label{fig:downstream_maps}
\end{figure}

\subsection{Labelled downstream task data} 

\subsubsection{Forest semantic and instance segmentation}

We utilize FOR-instanceV2 dataset~\citep{xiang2025forestformer3d}, \Cref{fig:downstream_maps} (a), a large-scale forest dataset that extends the earlier FOR-instance dataset~\citep{puliti2023_forinstance_paper}. It provides unique individual tree IDs for each point, as well as semantic labels for ground, wood, and leaf. The data were collected across multiple countries, including Norway, the Czech Republic, Austria, French Guiana, New Zealand, and Australia, and encompass ULS, MLS, TLS point cloud modalities. Overall, this dataset comprises 48 training plots, 17 validation plots, and 29 test plots, totalling 11,028 tree instances.

While this dataset provides a sufficiently large and diverse set of labelled forest plots, the annotation process requires substantial human efforts. In real-world forestry applications, access to large-scale annotated data is often limited. In contrast, this work focuses on developing a forest point cloud foundation model that can be effectively adapted to downstream applications with only a small amount of labelled data. To better reflect this practical setting, we partition the training data into multiple subsets with varying proportions (5\%, 10\%, 20\%, 50\%, 100\%) and report performance across these regimes. This protocol enables a more comprehensive assessment of model generalization and data efficiency, and establishes a more suitable benchmark to evaluate forest point cloud foundation models.

\subsubsection{Tree species classification}

For this task, where tree species labels are assigned to (segmented) scans of individual trees, we use the FOR-species20K~\citep{puliti2025_forspecies20k}, \Cref{fig:downstream_maps} (b). This dataset is primarily collected in Europe, with additional samples from Canada, Australia, and New Zealand, and covers TLS, ULS, and MLS sensor modalities. 
FOR-species20K contains 33 species, with \textit{Pinus sylvestris} and \textit{Fagus sylvatica} being the most common, while others such as \textit{Populus deltoides}, \textit{Corylus avellana}, \textit{Prunus avium} are underrepresented.
This dataset is highly imbalanced and its distribution reflects realistic species abundance in European forest ecosystems, where dominant tree species are prevalent and rarer species occur less frequently.
The full dataset is split into a development set (90\%, 17,707 trees) and test set (10\%, 2,255 trees). The development set is further divided into 14,165 trees for training and 3,542 trees for validation.

FOR-species20K represents the most comprehensive and largest dataset regarding the number of tree species openly available to date, providing a strong benchmark for developing and evaluating robust classification models. 
However, similar to the segmentation setting, real-world applications often lack access to large-scale data. To better assess the ability of foundation models to adapt under limited supervision, we partition the training set into subsets with varying proportions (1\%, 5\%, 10\%, 20\%, 50\%, 100\%) and conduct a comprehensive evaluation across these regimes.

\subsubsection{Tree age estimation}

For single-tree age regression, we use the FOR-age dataset~\citep{PULITI2026115462}, \Cref{fig:downstream_maps} (c). The dataset is collected from Norway and Finland, covering MLS, TLS, and ALSHD sensor modalities. In total, it comprises 1,775 tree scans spanning an age range of up to 350 years, with an average of 53 years. The dataset exhibits a long-tailed distribution, including seedling and saplings (6\% of trees younger than 10 years), established forests (50\% between 10 and 50 years), trees at typical harvesting maturity (28\% between 50 and 100 years), older trees (9\% between 100 and 200 years) and a small portion of very old trees (2\% older than 200 years). The dataset is split into 1,250 trees for training, 271 trees for validation, and 254 trees for testing. Similar with species classification, we split the training set into subsets with varying proportions (5\%, 10\%, 20\%, 50\%, 100\%) for data efficiency experiments.

\section{Methodology}
\label{sec_method}

\subsection{Overview}
We investigate two pretraining strategies, self-supervised and supervised pretraining, to assess their ability to learn transferable representations for forest point clouds and their potential as a step toward a 3D forest foundation model. 
For both strategies, a common LitePT backbone is first pretrained and then transferred to downstream forestry tasks through either frozen feature transfer or end-to-end fine-tuning. For comparison, we also train the same backbone from scratch on each downstream task. This unified framework enables a systematic comparison of alternative representation learning strategies under identical transfer learning protocols and annotation budgets.
The following sections describe the two pretraining strategies (\Cref{method_sub:ssl,method_sub:supervised}), the transfer learning protocol (\Cref{sub:transfer_learning}) and the pretraining implementation (\Cref{sub:implementation}).

\subsection{Self-supervised pretraining}
\label{method_sub:ssl}

\begin{figure}[!t]
\centering
\includegraphics[width=0.9\linewidth]{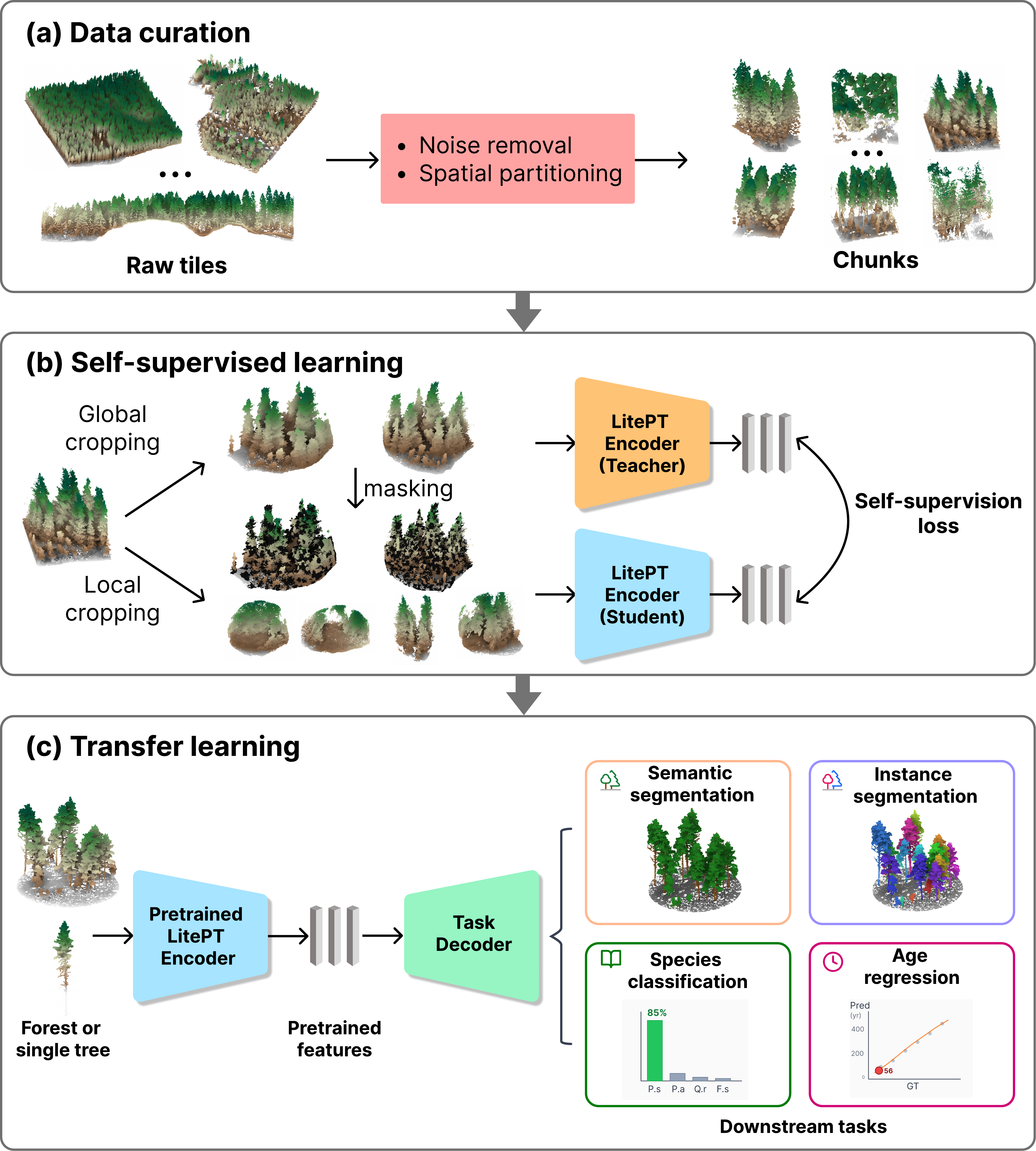}
\caption{\textbf{Self-supervised pretraining workflow.} A large-scale unlabelled forest point cloud corpus is curated and preprocessed. The LitePT backbone is then pretrained using teacher-student self-distillation. Finally, the pretrained encoder is applied to downstream forestry tasks through transfer learning.}\label{fig:ssl_pretraining_workflow}
\end{figure}

The overall self-supervised pretraining workflow is illustrated in ~\Cref{fig:ssl_pretraining_workflow}. First, we curate a large-scale unlabelled forest point cloud dataset and perform data preprocessing and standardization to obtain a unified representation suitable for neural network training. Next, the LitePT backbone is pretrained using self-supervised learning. Finally, the pretrained encoder is transferred to downstream forestry tasks using the transfer learning protocol described in~\Cref{sub:transfer_learning}.

\begin{figure}[ht]%
\centering%
\includegraphics[width=\linewidth]{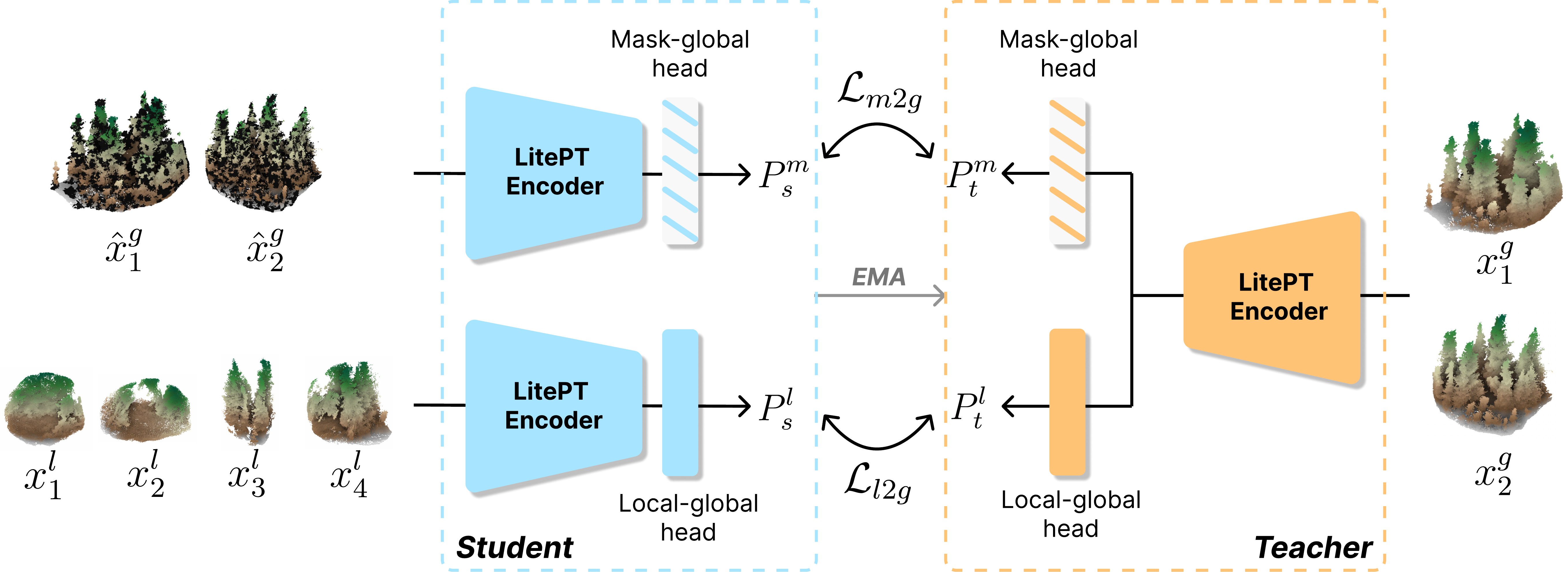}
\caption{\textbf{Self-supervised pretraining pipeline.} The framework follows a teacher-student architecture, where a student network is trained to match the outputs of a momentum-updated teacher network.}\label{fig:ssl_pipeline}
\end{figure}

The self-supervised learning architecture is shown in~\Cref{fig:ssl_pipeline}. We adopt a self-distillation framework inspired by DINOv2~\citep{oquab2023dinov2} and its point cloud adaptation Sonata~\citep{wu2025sonata}, and tailor it to the structural characteristics of forest point clouds. 
The overall architecture follows a student-teacher paradigm, where the student network is trained to match the output of a momentum-updated teacher network under multiple geometric views of the same input point cloud.

\noindent\textbf{Multi-view augmentation.} Given a forest point cloud pre-chunked into $20~\mathrm{m} \times 20~\mathrm{m}$ spatial tiles, we construct multiple stochastic views for self-supervised learning. Formally, for an input forest chunk $x$, we generate a set $\mathcal{V}$ of different views, consisting of two global views with large spatial support, $\mathcal{V}_g = \{ x_i^g \}_{i=1}^{2}$ and four local views with smaller crops, $\mathcal{V}_l = \{ x_i^l \}_{i=1}^{4}$. 
The global views retain most of the spatial extent of the original chunk, while the local views focus on partial regions, simulating incomplete observations commonly encountered in forest point clouds due to occlusion and varying sampling density.
All views are independently augmented using a set of geometric transformations, including random scaling, rotation, flipping, point jittering, and elastic distortion. These augmentations aim to enforce invariance to acquisition conditions and geometric perturbations while preserving the underlying forest semantics.
In addition, random patch masking~\citep{he2022masked,zhou2021ibot} is applied to the two global views to further increase task difficulty and promote contextual reasoning. Specifically, a subset of spatial patches is randomly removed from each global view, resulting in two masked global views, $\mathcal{\hat{V}}_g = \{ \hat{x}_i^g\}_{i=1}^{2}$. The masking strategy encourages the model to infer missing structure from partial observations, improving robustness to sparsity and occlusion commonly observed in LiDAR data.

\noindent\textbf{Network architecture.} The framework adopts a student-teacher architecture. 
Let $f_{\theta_s}$ and $f_{\theta_t}$ denote the student and teacher networks with parameters $\theta_s$ and $\theta_t$, respectively. Both networks share the same architecture, consisting of a backbone $b(\cdot)$ and two projection heads: a mask-global head $h^{m}(\cdot)$ and a local-global head $h^{l}(\cdot)$. We employ LitePT~\citep{yuelitept2026} as an efficient, scalable, and accurate backbone for forest point cloud learning. Each projection head is implemented as a three-layer multi-layer perceptron (MLP) followed by $\ell_2$ normalization and a weight-normalized fully connected layer. The student and teacher networks are initialized with identical weights. During training, gradients are not propagated through the teacher network, whose parameters are instead updated as an exponential moving average (EMA) of the student parameters:
\begin{equation}
\label{eq_ema}
\theta_t \leftarrow \tau \theta_t + (1 - \tau)\theta_s,
\end{equation}
where $\tau \in [0,1)$ is a momentum coefficient controlling the update rate.

\noindent\textbf{Self-distillation loss.} The training objective consists of two complementary branches: local-to-global distillation, and masked-to-global distillation. In the local-to-global distillation branch, the teacher backbone takes global views $x_i^g \in \mathcal{V}_g$ as input and the local-global head $h^{l}(\cdot)$ produces the target distribution $P^l_t$.
The student backbone processes the local views $x_i^l \in \mathcal{V}_l$ and the local-global head produces $P^l_s$. The local-to-global distillation loss is then defined as:
\begin{equation}
\mathcal{L}_{\mathrm{l2g}}
=
\sum_{x_i^l \in \mathcal{V}_l}
\sum_{x_j^g \in \mathcal{V}_g}
H\!\left(
P_t^{l}(x_j^g),
P_s^{l}(x_i^l)
\right).
\end{equation}
where $H(\cdot,\cdot)$ denotes the cross-entropy between the teacher and student distributions. 

In the masked-to-global distillation branch, the teacher backbone takes the \emph{unmasked} global views $x_i^g \in \mathcal{V}_g$ as input and the mask-global head $h^{m}(\cdot)$ produces $P^m_t$.
The student processes the corresponding masked global views $\hat{x}_i^g \in \mathcal{\hat{V}}_g$ and produces $P^m_s$. In addition to the aligned teacher-student pairs, $x_1^g \to \hat{x}_1^g$ and $x_2^g \to \hat{x}_2^g$, the loss also includes rolled cross-view pairs, $x_1^g \to \hat{x}_2^g$ and $x_2^g \to \hat{x}_1^g$. Thus, each masked global view is distilled from both teacher global views, and the loss is computed over all teacher-student global-view combinations:
\begin{equation}
\mathcal{L}_{\mathrm{m2g}}
=
\sum_{x_i^g \in \mathcal{V}_g}
\sum_{\hat{x}_j^g \in \hat{\mathcal{V}}_g}
H\!\left(
P_t^{m}(x_i^g),
P_s^{m}(\hat{x}_j^g)
\right).
\end{equation}
The final training objective is given by:
\begin{equation}
\label{eq_total_loss}
\mathcal{L} = \mathcal{L}_{\text{l2g}} + \mathcal{L}_{\text{m2g}}.
\end{equation}

\subsection{Supervised pretraining}
\label{method_sub:supervised}

\begin{figure}[!t]
\centering
\includegraphics[width=0.9\linewidth]{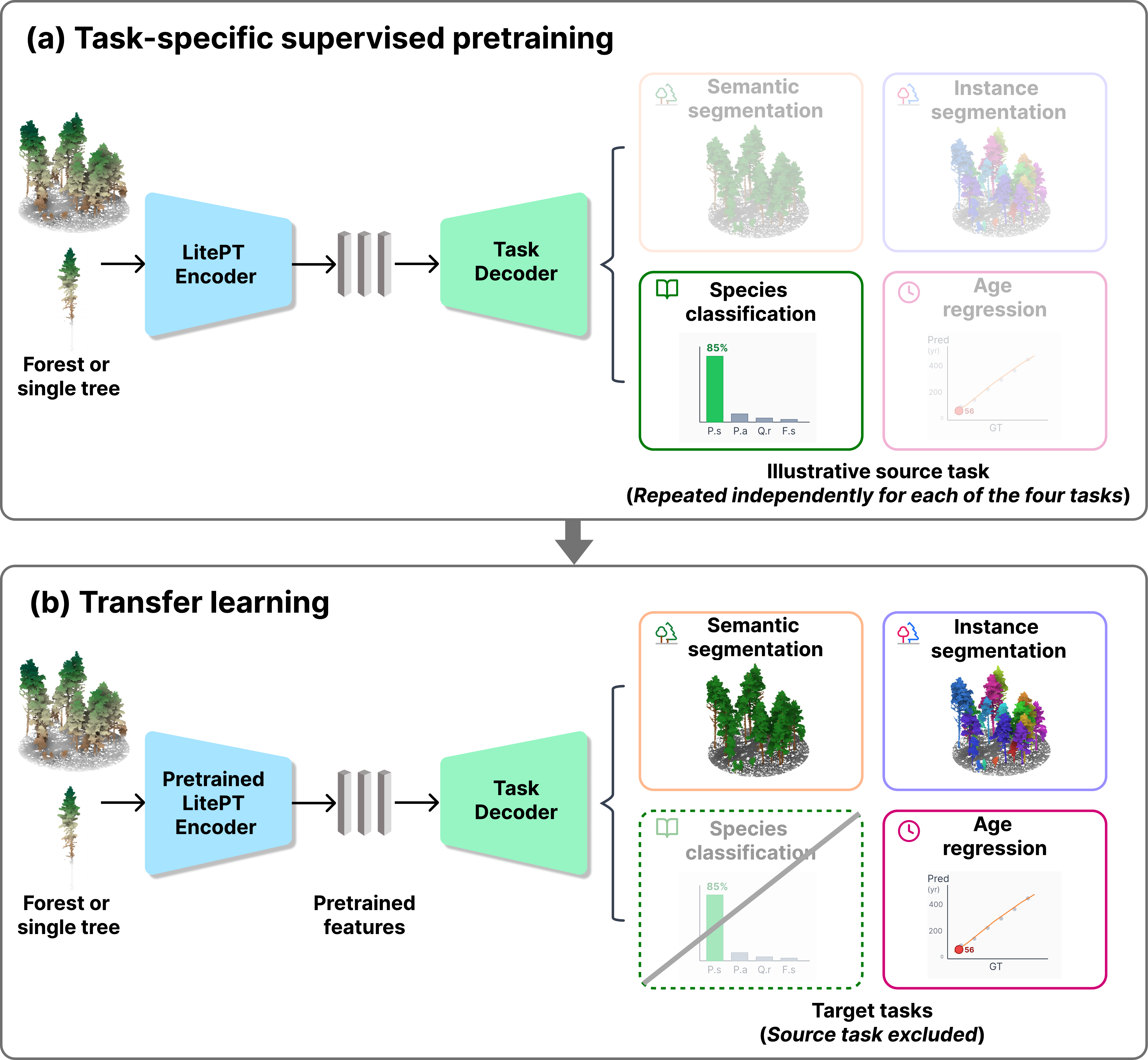}
\caption{\textbf{Task-specific supervised pretraining workflow.} An illustrative source task (species classification) is shown. The LitePT backbone is pretrained using full supervision for the source task, and the procedure is repeated independently for each of the four forestry tasks to obtain a separate pretrained encoder. The pretrained encoder is then transferred to the remaining downstream tasks; the source task is excluded from evaluation.}\label{fig:supervised_pretraining_workflow}
\end{figure}

The supervised pretraining workflow is illustrated in~\Cref{fig:supervised_pretraining_workflow}. Unlike self-supervised pretraining, which learns a single pretrained backbone from unlabelled data, supervised pretraining relies on task annotations and therefore produces a separate pretrained model for each source task. Each pretrained backbone is subsequently transferred to the remaining downstream tasks.

Specifically, we pretrain the same LitePT backbone independently on each of the four downstream tasks: forest semantic segmentation, forest instance segmentation, tree species classification, and tree age regression, using the same training procedures as the corresponding training-from-scratch baseline. After pretraining, the task-specific decoder or head is discarded, and the pretrained encoder is evaluated on the unseen downstream tasks.

As in forest instance segmentation, we employ the ForestFormer3D~\citep{xiang2025forestformer3d} head, which jointly optimizes semantic and instance predictions. Consequently, transfer between these two tasks would not constitute an independent evaluation of representation transfer. We therefore exclude semantic-to-instance and instance-to-semantic transfer from our experiments.

\subsection{Transfer learning}
\label{sub:transfer_learning}

Following pretraining, the pretrained LitePT encoder is transferred to downstream forestry tasks using either frozen probing or end-to-end fine-tuning.

\noindent\textbf{Frozen probing.} To directly leverage the learned representations with minimal task-specific adaptation, we freeze the pretrained encoder and optimize only a lightweight task-specific prediction module. Depending on the downstream architecture, this prediction module may consist of a linear layer, a LitePT decoder, or a task-specific prediction head. 
The specific probing configurations vary across downstream tasks and are described in \Cref{sec_experiments}.

\noindent\textbf{End-to-end fine-tuning.} To fully adapt the pretrained features to a downstream task, we initialize the LitePT encoder with pretrained weights and jointly optimize the entire model with the task-specific decoder or prediction head. The fine-tuning architecture is identical to that used for the corresponding training-from-scratch baseline. 

\noindent\textbf{Partial weights loading.}
Besides initializing the entire encoder, we also investigate partial weight initialization, where only the first $k$ stages of the pretrained LitePT encoder are loaded while the remaining stages are randomly initialized. This setting allows us to study how representations learned at different levels of the encoder affect the downstream transfer. 

\subsection{Backbone and pretraining implementation} 
\label{sub:implementation}

We use LitePT~\citep{yuelitept2026} as our backbone across all experiments.
Following the recommendations of \citet{wu2025sonata}, we make the following modifications to improve its compatibility with self-supervised learning. 
First, we replace the Batch Normalization~\citep{ioffe2015batch} layers in LitePT with Layer Normalization~\citep{ba2016layer} to improve training stability and enhance generalization to unseen datasets. 
Second, we change the embedding layer from a single sparse convolution layer to a single linear layer. Lastly, LitePT adopts an encoder-decoder U-Net architecture, while pretraining is only performed on the encoder. 

Self-supervised pretraining is conducted on 64 NVIDIA GH200 GPUs with a total batch size of 128 for 30 epochs. We use a constant learning rate of 0.001, a weight decay of 0.0001, and a layer-wise learning rate decay factor of 0.9. The mask size is set to 5 cm with a mask ratio of 0.7, and grid sampling is performed using a 5 cm grid size. For EMA training, the student and teacher temperatures are set to 0.1 and 0.07, respectively, while the momentum is fixed at 0.994.

Unless otherwise specified, supervised pretraining adopts the same optimization settings as the corresponding training-from-scratch baselines. Implementation details specific to each downstream task are provided in \ref{app_implementation}.

\section{Experiments}
\label{sec_experiments}

\begin{figure}[!t]%
\centering%
\includegraphics[width=0.9\linewidth]{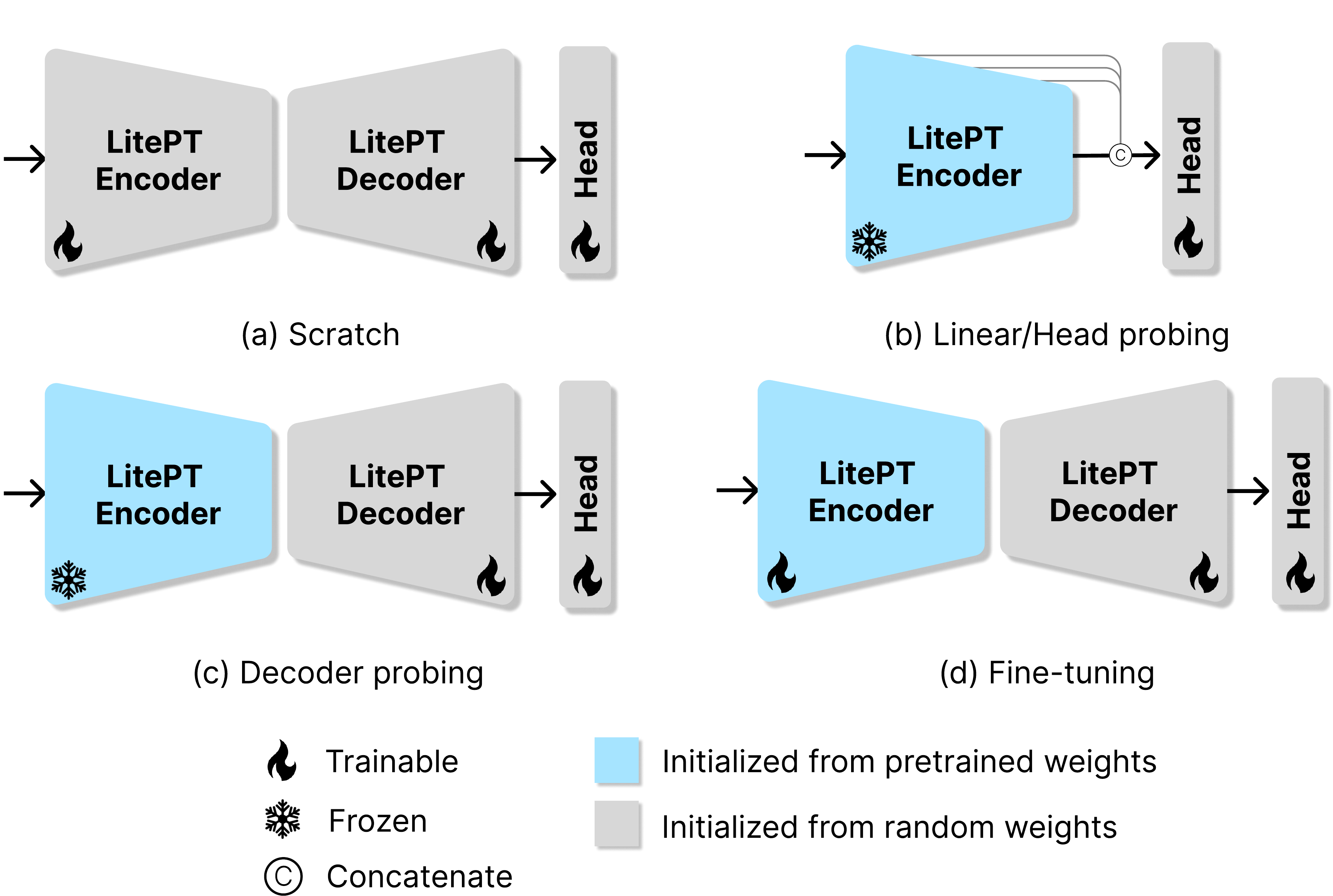}
\caption{\textbf{Model variants for forest semantic segmentation and instance segmentation.} For semantic segmentation, the task head is a single linear layer. For instance segmentation, the task head is the ForestFormer3D~\citep{xiang2025forestformer3d} decoder.}\label{fig:semantic_instance_model_variants}
\end{figure}

\subsection{Overall setup}
We consider four forestry downstream tasks: forest semantic segmentation, forest instance segmentation, tree species classification, and tree age regression.  
Our experiments address two complementary questions. First, we investigate whether self-supervised pretraining improves downstream performance over training from scratch (\Cref{subsec:sem,subsec:ins,subsec:cls,subsec:age}). Second, we evaluate whether task-specific supervised pretraining learns transferable representations (\Cref{subsec:supervised_transfer}).

Unless otherwise specified, experiments are conducted under multiple annotation budgets to simulate different levels of supervision. The same annotation subsets are used across all methods to ensure fair comparisons. Depending on the downstream task, we consider training from scratch, frozen probing, end-to-end fine-tuning, and where applicable, $k$NN evaluation. Since dense prediction and global prediction tasks adopt different downstream architectures, the exact probing configurations are described in the corresponding task subsections.

\subsection{Semantic segmentation}
\label{subsec:sem}

\noindent\textbf{Setup.} In this task, we train a model to assign each point in a forest point cloud to three classes: wood, leaf, and ground. The different model variants are illustrated in~\Cref{fig:semantic_instance_model_variants}. We employ a single linear layer as the segmentation head. As is standard for dense prediction, we employ a LitePT decoder to aggregate multi-scale features before the segmentation head, except in the linear probing setting, where the segmentation head operates directly on concatenated multi-scale encoder features.

In the \emph{scratch} setting, the encoder, decoder, and segmentation head are randomly initialized and trained from scratch.
In the \emph{linear probing} setting, we freeze the pretrained encoder and train only the segmentation head.
In the \emph{decoder probing} setting, we keep the pretrained encoder frozen while randomly initializing and jointly training the decoder and segmentation head.
In the \emph{fine-tuning} setting, we initialize the encoder with pretrained weights, randomly initialize the decoder and segmentation head, and jointly optimize all modules.

In addition to learned evaluation, we employ a non-parametric segmentation $k$-nearest neighbour ($k$NN) label-transfer protocol to directly measures semantic class separability in the learned feature space. To keep nearest-neighbour search tractable at the scale of forest point clouds, we operated on geometric superpoints~\citep{robert2023efficient, geist2026ezsp} rather than individual points. Each superpoint is represented by the mean of its constituent point features and labelled by majority vote over its full-resolution ground-truth labels. Superpoints extracted from the training scenes constitute a labelled reference bank. Each test superpoint is assigned the majority label of its $k$ nearest bank superpoints under cosine similarity in feature space, and predictions are propagated to full resolution via the point-to-superpoint map. We use $k=8$ throughout the experiments.

\noindent\textbf{Evaluation metrics.} 
Following standard semantic segmentation evaluation protocols, we report mean Intersection over Union (mIoU), mean class accuracy (mAcc), and overall accuracy (allAcc), which are defined as:

\begin{equation}
\mathrm{IoU}_c = \frac{\mathrm{intersection}_c}{\mathrm{union}_c},
\end{equation}

\begin{equation}
\mathrm{mIoU} = \frac{1}{C}\sum_{c=1}^{C}\mathrm{IoU}_c,
\end{equation}

\begin{equation}
\mathrm{Acc}_c = \frac{\mathrm{intersection}_c}{\mathrm{target}_c},
\end{equation}

\begin{equation}
\mathrm{mAcc} = \frac{1}{C}\sum_{c=1}^{C}\mathrm{Acc}_c,
\label{eq_macc}
\end{equation}

\begin{equation}
\mathrm{allAcc} = \frac{\sum_{c=1}^{C}\mathrm{intersection}_c}{\sum_{c=1}^{C}\mathrm{target}_c},
\label{all_macc}
\end{equation}

where $\mathrm{intersection}_c$ denotes the number of correctly predicted points for class $c$, $\mathrm{union}_c$ denotes the union of predicted and ground-truth points for class $c$, $\mathrm{target}_c$ denotes the number of ground-truth points belonging to class $c$, and $C$ is the total number of semantic classes. mIoU evaluates the segmentation performance across semantic classes, mAcc measures the averaged class-wise prediction accuracy, and allAcc evaluates the overall point-wise classification accuracy across the entire evaluation set.

\begin{table}[ht]
\centering
\setlength{\tabcolsep}{2pt}
\resizebox{\columnwidth}{!}{
\begin{tabular}{clcccccc}
\toprule
\cmidrule{1-8} 
&  & \multicolumn{3}{c}{Overall} & \multicolumn{3}{c}{Per-class IoU} \\
\cmidrule(r){3-5}       \cmidrule(r){6-8}
Training Data & Setup~{\scriptsize(trainable/total parameters)} &   mIoU (\%) $\uparrow$ & mAcc (\%) $\uparrow$ & allAcc (\%) $\uparrow$ & Ground (\%) $\uparrow$ & Leaf (\%) $\uparrow$ & Wood (\%) $\uparrow$ \\
\midrule
\multirow{5}{*}{0.15ha (5\%)} & 
Scratch~{\scriptsize(16.0M/16.0M)} & \cellcolor{rankgreen!0}\textcolor{black!65}{57.9} & \cellcolor{rankgreen!0}\textcolor{black!65}{63.4} & \cellcolor{rankgreen!0}\textcolor{black!65}{85.7} & \cellcolor{rankgreen!18}\textcolor{black!65}{73.6} & \cellcolor{rankgreen!0}\textcolor{black!65}{85.3} & \cellcolor{rankgreen!0}\textcolor{black!65}{14.7} \\
& $k$NN~{\scriptsize(0/12.4M)} & \cellcolor{rankgreen!24}\textbf{68.5} & \cellcolor{rankgreen!24}\textbf{72.8} & \cellcolor{rankgreen!6}89.3 & \cellcolor{rankgreen!24}\textbf{85.9} & \cellcolor{rankgreen!6}88.4 & \cellcolor{rankgreen!6}31.0  \\
& Linear Probing~{\scriptsize(3.0K/12.4M)} & \cellcolor{rankgreen!6}59.3 & \cellcolor{rankgreen!6}63.8 & \cellcolor{rankgreen!12}89.4 & \cellcolor{rankgreen!0}47.9 & \cellcolor{rankgreen!12}88.5 & \cellcolor{rankgreen!18}41.3  \\
& Decoder Probing~{\scriptsize(3.6M/16.0M)} & \cellcolor{rankgreen!12}61.2 & \cellcolor{rankgreen!12}65.5 & \cellcolor{rankgreen!18}89.7 & \cellcolor{rankgreen!12}54.2 & \cellcolor{rankgreen!18}88.9 & \cellcolor{rankgreen!12}40.5  \\
& Fine-tuning~{\scriptsize(16.0M/16.0M)} & \cellcolor{rankgreen!18}61.3 & \cellcolor{rankgreen!18}65.5 & \cellcolor{rankgreen!24}\textbf{90.1} & \cellcolor{rankgreen!6}50.9 & \cellcolor{rankgreen!24}\textbf{89.4} & \cellcolor{rankgreen!24}\textbf{43.6}  \\
\arrayrulecolor{black!10}\midrule\arrayrulecolor{black}
\multirow{5}{*}{0.34ha (10\%)} & 
Scratch & \cellcolor{rankgreen!12}\textcolor{black!65}{78.8} & \cellcolor{rankgreen!12}\textcolor{black!65}{86.2} & \cellcolor{rankgreen!6}\textcolor{black!65}{91.5} & \cellcolor{rankgreen!6}\textcolor{black!65}{93.4} & \cellcolor{rankgreen!6}\textcolor{black!65}{90.3} & \cellcolor{rankgreen!6}\textcolor{black!65}{52.8} \\
& $k$NN & \cellcolor{rankgreen!6}78.2 & \cellcolor{rankgreen!6}83.2 & \cellcolor{rankgreen!12}91.6 & \cellcolor{rankgreen!24}\textbf{95.1} & \cellcolor{rankgreen!12}90.5 & \cellcolor{rankgreen!0}49.0  \\
& Linear Probing & \cellcolor{rankgreen!0}69.3 & \cellcolor{rankgreen!0}75.7 & \cellcolor{rankgreen!0}91.1 & \cellcolor{rankgreen!0}62.2 & \cellcolor{rankgreen!0}90.1 & \cellcolor{rankgreen!12}55.5  \\
& Decoder Probing & \cellcolor{rankgreen!18}81.6 & \cellcolor{rankgreen!24}\textbf{87.0} & \cellcolor{rankgreen!18}93.2 & \cellcolor{rankgreen!12}93.5 & \cellcolor{rankgreen!18}92.3 & \cellcolor{rankgreen!18}58.9  \\
& Fine-tuning & \cellcolor{rankgreen!24}\textbf{82.2} & \cellcolor{rankgreen!18}86.5 & \cellcolor{rankgreen!24}\textbf{93.8} & \cellcolor{rankgreen!18}94.0 & \cellcolor{rankgreen!24}\textbf{93.0} & \cellcolor{rankgreen!24}\textbf{59.6}  \\
\arrayrulecolor{black!10}\midrule\arrayrulecolor{black}
\multirow{5}{*}{0.81ha (20\%)} & 
Scratch & \cellcolor{rankgreen!18}\textcolor{black!65}{83.3} & \cellcolor{rankgreen!18}\textcolor{black!65}{88.1} & \cellcolor{rankgreen!18}\textcolor{black!65}{93.9} & \cellcolor{rankgreen!12}\textcolor{black!65}{95.2} & \cellcolor{rankgreen!18}\textcolor{black!65}{93.1} & \cellcolor{rankgreen!12}\textcolor{black!65}{61.7} \\
& $k$NN & \cellcolor{rankgreen!6}80.5 & \cellcolor{rankgreen!6}85.8 & \cellcolor{rankgreen!6}92.4 & \cellcolor{rankgreen!24}\textbf{96.3} & \cellcolor{rankgreen!6}91.3 & \cellcolor{rankgreen!0}54.1  \\
& Linear Probing & \cellcolor{rankgreen!0}74.3 & \cellcolor{rankgreen!0}80.8 & \cellcolor{rankgreen!0}91.9 & \cellcolor{rankgreen!0}74.8 & \cellcolor{rankgreen!0}91.0 & \cellcolor{rankgreen!6}57.2  \\
& Decoder Probing & \cellcolor{rankgreen!12}82.6 & \cellcolor{rankgreen!12}87.7 & \cellcolor{rankgreen!12}93.9 & \cellcolor{rankgreen!6}92.8 & \cellcolor{rankgreen!12}93.0 & \cellcolor{rankgreen!18}62.1  \\
& Fine-tuning & \cellcolor{rankgreen!24}\textbf{85.3} & \cellcolor{rankgreen!24}\textbf{89.5} & \cellcolor{rankgreen!24}\textbf{94.8} & \cellcolor{rankgreen!18}95.3 & \cellcolor{rankgreen!24}\textbf{94.0} & \cellcolor{rankgreen!24}\textbf{66.5}  \\
\arrayrulecolor{black!10}\midrule\arrayrulecolor{black}
\multirow{5}{*}{2.59ha (50\%)} & 
Scratch & \cellcolor{rankgreen!18}\textcolor{black!65}{86.2} & \cellcolor{rankgreen!18}\textcolor{black!65}{90.9} & \cellcolor{rankgreen!18}\textcolor{black!65}{94.9} & \cellcolor{rankgreen!24}\textcolor{black!65}{\textbf{97.3}} & \cellcolor{rankgreen!18}\textcolor{black!65}{94.0} & \cellcolor{rankgreen!18}\textcolor{black!65}{67.1} \\
& $k$NN & \cellcolor{rankgreen!6}81.5 & \cellcolor{rankgreen!6}87.2 & \cellcolor{rankgreen!0}92.7 & \cellcolor{rankgreen!18}96.2 & \cellcolor{rankgreen!0}91.6 & \cellcolor{rankgreen!0}56.7  \\
& Linear Probing & \cellcolor{rankgreen!0}80.7 & \cellcolor{rankgreen!0}86.7 & \cellcolor{rankgreen!6}93.0 & \cellcolor{rankgreen!0}91.3 & \cellcolor{rankgreen!6}92.0 & \cellcolor{rankgreen!6}58.7  \\
& Decoder Probing & \cellcolor{rankgreen!12}84.2 & \cellcolor{rankgreen!12}89.7 & \cellcolor{rankgreen!12}94.3 & \cellcolor{rankgreen!6}93.9 & \cellcolor{rankgreen!12}93.4 & \cellcolor{rankgreen!12}65.3  \\
& Fine-tuning & \cellcolor{rankgreen!24}\textbf{86.5} & \cellcolor{rankgreen!24}\textbf{91.1} & \cellcolor{rankgreen!24}\textbf{95.2} & \cellcolor{rankgreen!12}95.8 & \cellcolor{rankgreen!24}\textbf{94.4} & \cellcolor{rankgreen!24}\textbf{69.4}  \\
\arrayrulecolor{black!10}\midrule\arrayrulecolor{black}
\multirow{5}{*}{5.71ha (100\%)} & 
Scratch & \cellcolor{rankgreen!12}\textcolor{black!65}{87.2} & \cellcolor{rankgreen!18}\textcolor{black!65}{92.1} & \cellcolor{rankgreen!12}\textcolor{black!65}{95.1} & \cellcolor{rankgreen!6}\textcolor{black!65}{98.4} & \cellcolor{rankgreen!12}\textcolor{black!65}{94.3} & \cellcolor{rankgreen!18}\textcolor{black!65}{68.9} \\
& $k$NN & \cellcolor{rankgreen!0}83.2 & \cellcolor{rankgreen!0}88.4 & \cellcolor{rankgreen!0}93.3 & \cellcolor{rankgreen!12}98.5 & \cellcolor{rankgreen!0}92.3 & \cellcolor{rankgreen!0}59.0  \\
& Linear Probing & \cellcolor{rankgreen!6}84.2 & \cellcolor{rankgreen!6}89.1 & \cellcolor{rankgreen!6}94.0 & \cellcolor{rankgreen!0}98.0 & \cellcolor{rankgreen!6}93.0 & \cellcolor{rankgreen!6}61.5  \\
& Decoder Probing & \cellcolor{rankgreen!18}87.3 & \cellcolor{rankgreen!12}91.5 & \cellcolor{rankgreen!18}95.2 & \cellcolor{rankgreen!18}98.9 & \cellcolor{rankgreen!18}94.4 & \cellcolor{rankgreen!12}68.5  \\
& Fine-tuning & \cellcolor{rankgreen!24}\textbf{88.3} & \cellcolor{rankgreen!24}\textbf{92.4} & \cellcolor{rankgreen!24}\textbf{95.6} & \cellcolor{rankgreen!24}\textbf{99.1} & \cellcolor{rankgreen!24}\textbf{94.9} & \cellcolor{rankgreen!24}\textbf{71.0}  \\
\bottomrule
\end{tabular}
}
\caption{
\textbf{Forest semantic segmentation results on FOR-instanceV2 test set.} Comparison of training from scratch and transfer from the self-supervised pretrained backbone. The data percentage is computed based on the number of training samples and is not directly proportional to the covered area. Values in parentheses (shown once in the top block) denote each setup's trainable / total parameters, which are identical across data fractions.
}
\label{tab:sem_seg}
\end{table}

\begin{figure}[!t]%
\centering%
\includegraphics[width=0.5\linewidth]{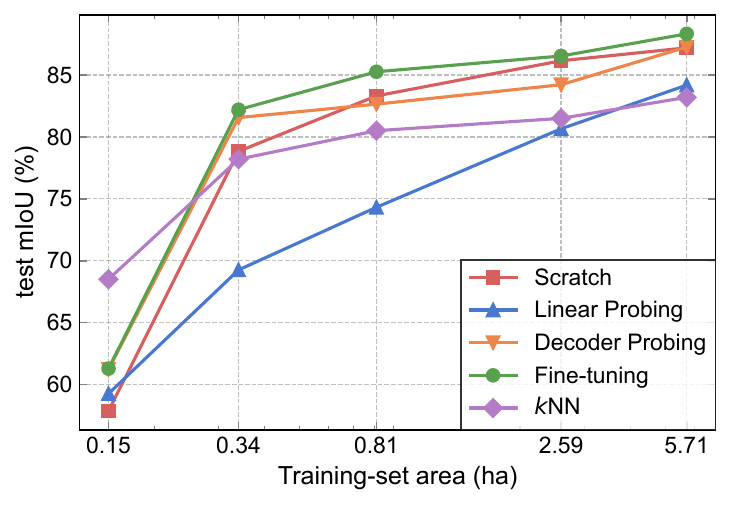}
\caption{\textbf{Data efficiency of semantic segmentation under different adaptation strategies.} Comparison of training from scratch and transfer from the self-supervised pretrained backbone across five training data fractions on a logarithmic scale. Self-supervised pretraining provides the largest gains in the low-data regime.}\label{fig:sem_data_efficiency}
\end{figure}

\noindent\textbf{Results.} \Cref{tab:sem_seg} summarizes the forest semantic segmentation performance on the FOR-instanceV2 test set under varying proportions of labelled training data (5\%-100\%). Four main observations can be drawn. First, fine-tuning consistently outperforms training from scratch across all data regimes, demonstrating the effectiveness of semantic features learned during pretraining. \
Second, the benefits of leveraging pretrained features is most pronounced in low-data regimes and gradually diminishes as the amount of labelled data increases, as further illustrated in \Cref{fig:sem_data_efficiency}. 
This behavior is particularly desirable in practical applications, where labelled data is often scarce and pretraining provides the greatest benefit. 
Third, in low-data settings (e.g., with only 0.15ha), the non-parametric $k$NN probe outperforms all parametric baselines and remains competitive even when the full 5.71ha training data is used. This pattern is coherent: when labels are scarce, $k$NN has no parameters to overfit and simply reads out the class structure already embedded in the frozen features.
Finally, in terms of performance per-class, the wood class remains the most challenging in all settings. Fine-tuning leads to the most significant improvements for this class, underscoring the importance of pretrained representations in capturing complex structural patterns.

\begin{figure}[!t]%
\centering%
\includegraphics[width=0.9\linewidth]{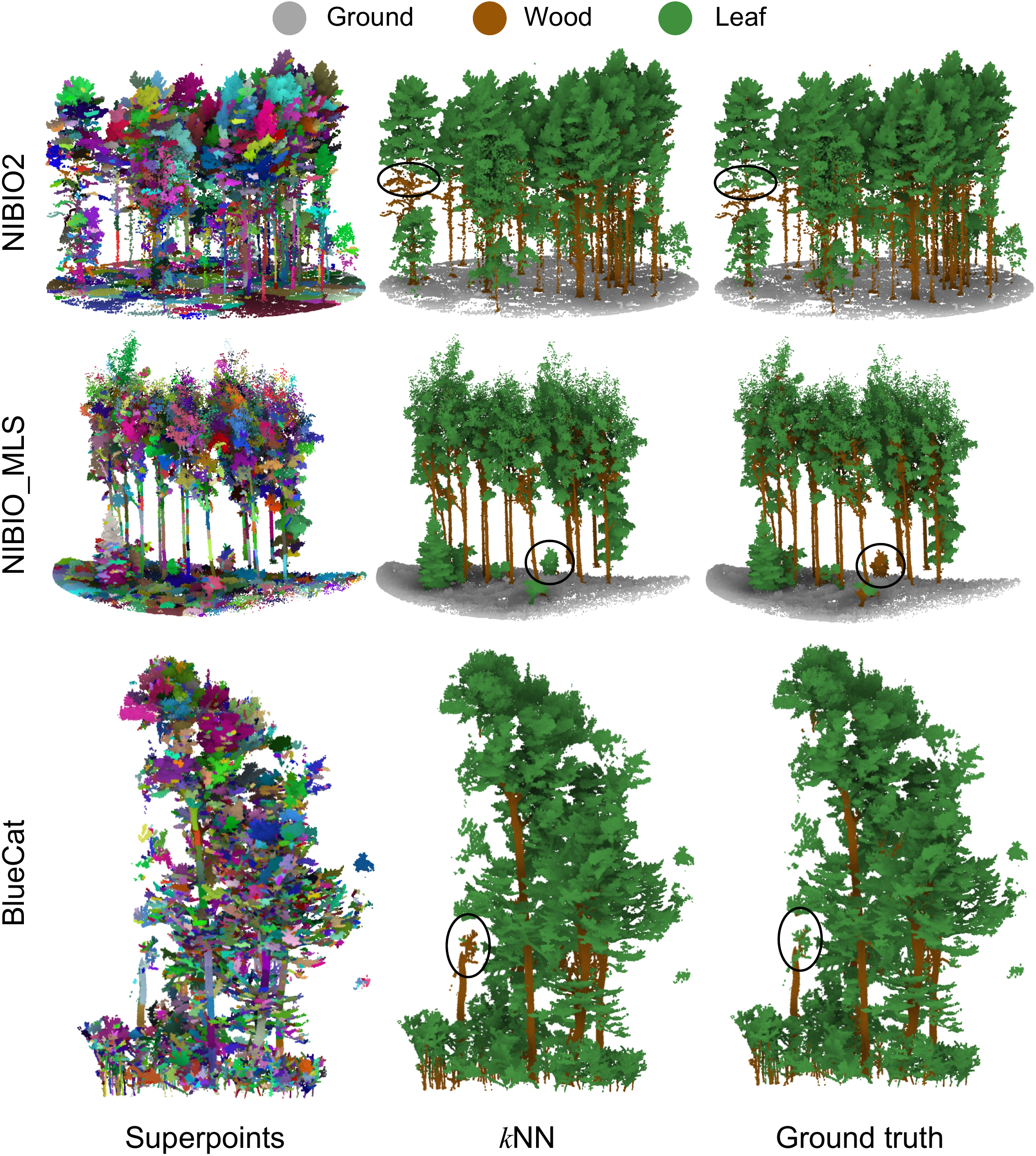}
\caption{\textbf{Qualitative forest semantic segmentation results on FOR-instanceV2 test set samples.} A non-parametric $k$NN probe is applied to representations learned by self-supervised pretraining.
}\label{fig:visuals_knn_sem}
\end{figure}

\Cref{fig:visuals_knn_sem} presents qualitative semantic segmentation results obtained using the $k$NN probe. The first column shows the superpoint partitioning. Using finer superpoints could further improve segmentation performance, albeit at the expense of increased computational cost. Despite this extremely simple, training-free protocol, the method already achieves strong semantic segmentation performance, indicating the learned feature space effectively separates semantic classes. The learned representation is robust in challenging environments where both large and small trees coexist. As shown in the last row, it delivers strong segmentation results on the BlueCat dataset, which was acquired using terrestrial laser scanning (TLS). TLS data were never used in pretraining, demonstrating that the learned representation generalizes well to point clouds acquired by unseen sensing modalities.

\begin{figure}[!t]%
\centering%
\includegraphics[width=\linewidth]{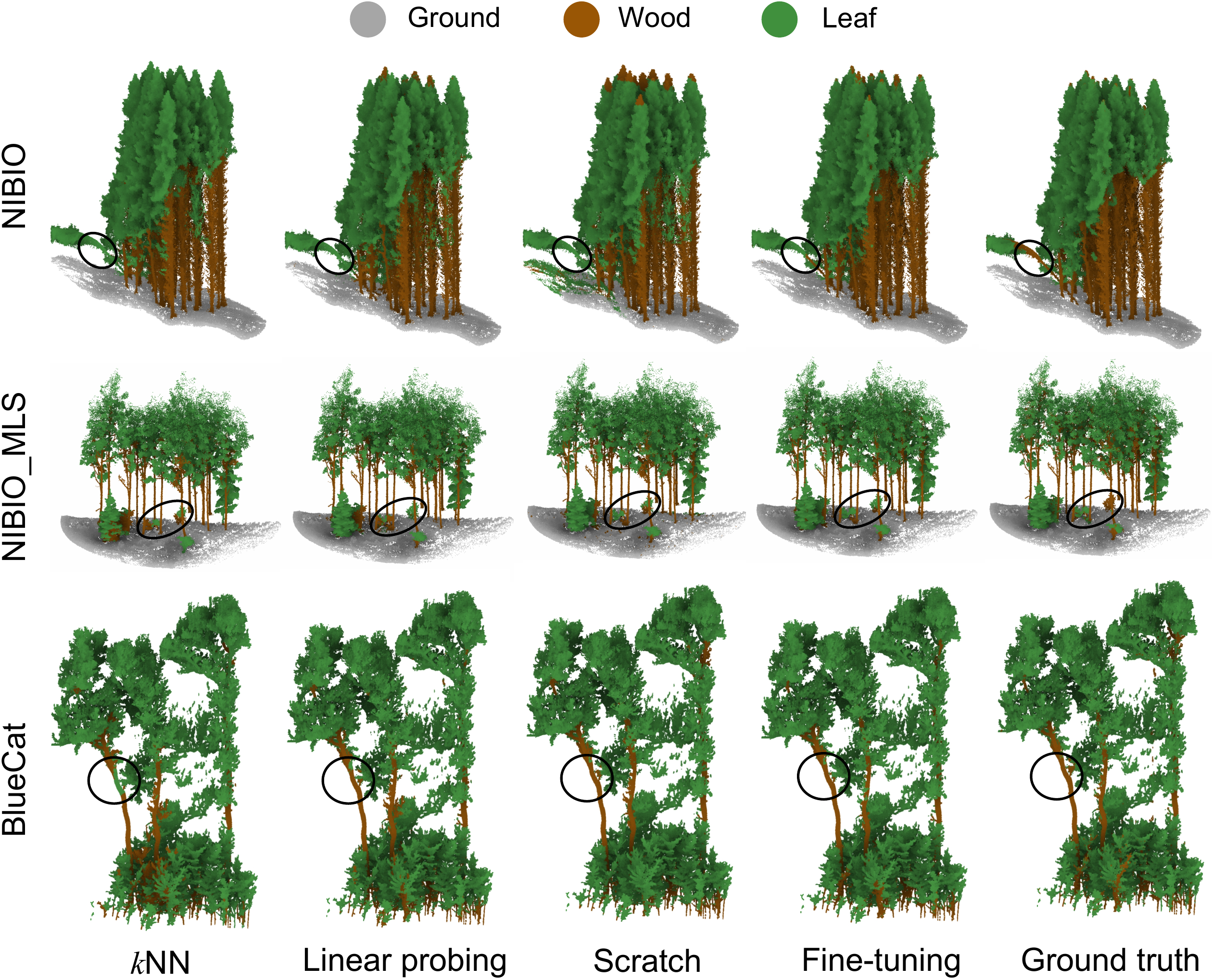}
\caption{\textbf{Qualitative forest semantic segmentation results on FOR-instanceV2 test set samples when 0.81ha labelled training data is used.} Comparison of training from scratch and transfer from the self-supervised pretrained backbone.}\label{fig:visuals_sem}
\end{figure}

We further compare the qualitative semantic segmentation results obtained with different training strategies using 0.81ha (20\%) labelled training data in \Cref{fig:visuals_sem}. The visualization includes forest scenes acquired by different sensing platforms: NIBIO (ULS), NIBO\_MLS (MLS), and BlueCat (TLS). The qualitative results are consistent with the quantitative trends reported in \Cref{tab:sem_seg}. Specifically, both simple $k$NN and linear probing produce reasonably strong segmentation results, while fine-tuning consistently outperforms training from scratch. For example, in the first row, the model trained from scratch fails to segment the stem of a bent tree, whereas the finetuned model produces a more accurate stem segmentation.

\begin{figure}[ht]%
\centering%
\includegraphics[width=\linewidth]{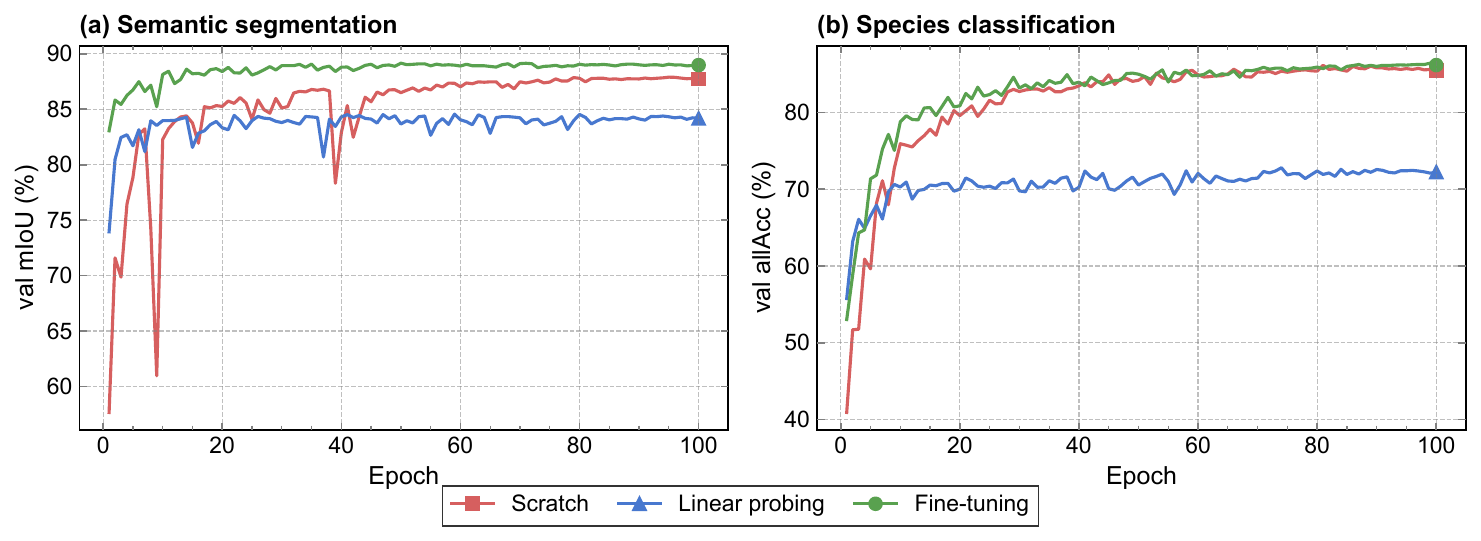}
\caption{\textbf{Validation performance over training epochs} for training from scratch, linear probing, and fine-tuning of the self-supervised pretrained backbone, on (a) semantic segmentation (mIoU, \%); (b) tree species classification (Overall accuracy, \%).}
\label{fig:val_curve}
\end{figure}

Beyond data efficiency, we also examine training efficiency. \Cref{fig:val_curve} (a) shows the validation mIoU over training epochs of the three training strategies: training from scratch, linear probing, and fine-tuning. Both fine-tuning and linear probing, which leverage pretrained representations, achieve strong performance early in training, demonstrating significantly faster convergence compared to training from scratch. In particular, fine-tuning consistently attains the highest mIoU throughout training and converges to the best final performance. Linear probing also benefits from pretraining, achieving competitive accuracy within only a few epochs, highlighting the effectiveness of the pretrained features in resource-constrained settings. In contrast, training from scratch requires substantially more epochs to achieve comparable performance and exhibits less stable optimization in the early stages. Overall, these results highlight the advantage of pretraining in accelerating convergence, improving stability, and enabling strong performance even with minimal training.

\subsection{Instance segmentation}
\label{subsec:ins}

\noindent\textbf{Setup.} In this task, given a forest point cloud as input, the goal is to segment individual trees by assigning each point to a unique instance ID corresponding to a specific tree, while treating the ground as background. We adopt ForestFormer3D~\citep{xiang2025forestformer3d},  the current state-of-the-art method that jointly performs semantic segmentation and individual tree segmentation. 

ForestFormer3D originally employs a Sparse U-Net~\citep{graham20183d,choy20194d} as backbone for feature extraction. The backbone features are passed to two parallel MLP heads: one learns instance discriminative features, and the other predicts tree versus non-tree classifications, which are used to sample instance queries. These sampled instance queries, together with learnable semantic queries are then processed by a Transformer-based query decoder to produce individual tree masks and semantic segmentation outputs. ForestFormer3D adopts a two-stage training strategy, where the two MLP heads are pretrained before jointly training the full model.

In our implementation, we replace the original backbone with LitePT, a high-performing and lightweight backbone. With this modification, we observe that directly training the full model end-to-end from the start yields better performance while simplifying the training pipeline, removing the need for the two-stage procedure. Additionally, for improved training efficiency, we pre-chunk two large forest scenes (BlueCat and Yuchen) to facilitate data sampling. During inference, we follow the protocol of ForestFormer3D by sampling overlapping cylindrical regions and merging predictions across blocks based on confidence scores. 

The training configurations are illustrated in ~\Cref{fig:semantic_instance_model_variants}.
In the \emph{scratch} setting, all weights in the LitePT-enhanced ForestFormer3D are randomly initialized. 
In the \emph{head probing} setting, we freeze the pretrained encoder and train only the ForestFormer3D prediction head from random initialization.
In the \emph{decoder probing} setting, we keep the pretrained encoder frozen, while randomly initializing and jointly training the decoder and the ForestFormer3D prediction head.
In the \emph{fine-tuning} setting, we initialize the encoder of LitePT with pretrained weights. Notably, instead of loading weights of all encoder stages, we find that initializing only the first two stages yields the best performance, and we adopt this strategy for this task. The decoder and ForestFormer3D prediction head are randomly initialized, and all modules are optimized jointly.

\begin{table}[!t]
\centering
\setlength{\tabcolsep}{20pt}
\resizebox{\columnwidth}{!}{
\begin{tabular}{clccc}
\toprule
\cmidrule{1-5} 
& & \multicolumn{3}{c}{Overall}  \\
 \cmidrule(r){3-5}
Training Data & Setup~{\scriptsize(trainable/total parameters)}  &   mAP $\uparrow$ &  mAP$_{25}$ $\uparrow$ & mAP$_{50}$ $\uparrow$ \\
\midrule
\multirow{4}{*}{0.29ha (5\%)} & Scratch~{\scriptsize(22.5M/22.5M)} & \cellcolor{rankgreen!16}\textcolor{black!65}{26.6} & \cellcolor{rankgreen!8}\textcolor{black!65}{54.3} & \cellcolor{rankgreen!16}\textcolor{black!65}{43.0} \\
 & Head Probing~{\scriptsize(7.4M/19.8M)} & \cellcolor{rankgreen!0}4.3 & \cellcolor{rankgreen!0}17.1 & \cellcolor{rankgreen!0}9.3 \\
 & Decoder Probing~{\scriptsize(10.1M/22.5M)} & \cellcolor{rankgreen!8}25.6 & \cellcolor{rankgreen!16}\textbf{55.6} & \cellcolor{rankgreen!8}42.5 \\
 & Fine-tuning~{\scriptsize(22.5M/22.5M)} & \cellcolor{rankgreen!24}\textbf{28.2} & \cellcolor{rankgreen!16}\textbf{55.6} & \cellcolor{rankgreen!24}\textbf{45.7} \\

\arrayrulecolor{black!10}\midrule\arrayrulecolor{black}
\multirow{4}{*}{0.55ha (10\%)} & Scratch & \cellcolor{rankgreen!16}\textcolor{black!65}{40.9} & \cellcolor{rankgreen!24}\textcolor{black!65}{\textbf{67.4}} & \cellcolor{rankgreen!16}\textcolor{black!65}{58.5} \\
 & Head Probing & \cellcolor{rankgreen!0}12.0 & \cellcolor{rankgreen!0}31.2 & \cellcolor{rankgreen!0}22.9 \\
 & Decoder Probing & \cellcolor{rankgreen!8}36.3 & \cellcolor{rankgreen!8}63.6 & \cellcolor{rankgreen!8}53.7 \\
 & Fine-tuning & \cellcolor{rankgreen!24}\textbf{41.1} & \cellcolor{rankgreen!16}66.6 & \cellcolor{rankgreen!24}\textbf{59.1} \\

\arrayrulecolor{black!10}\midrule\arrayrulecolor{black}
\multirow{4}{*}{1.35ha (20\%)} & Scratch & \cellcolor{rankgreen!16}\textcolor{black!65}{48.0} & \cellcolor{rankgreen!16}\textcolor{black!65}{71.5} & \cellcolor{rankgreen!16}\textcolor{black!65}{\textbf{65.7}} \\
 & Head Probing & \cellcolor{rankgreen!0}20.3 & \cellcolor{rankgreen!0}41.3 & \cellcolor{rankgreen!0}34.4 \\
 & Decoder Probing & \cellcolor{rankgreen!8}44.5 & \cellcolor{rankgreen!8}69.4 & \cellcolor{rankgreen!8}61.4 \\
 & Fine-tuning & \cellcolor{rankgreen!24}\textbf{48.2} & \cellcolor{rankgreen!24}\textbf{73.4} & \cellcolor{rankgreen!16}\textbf{65.7} \\

\arrayrulecolor{black!10}\midrule\arrayrulecolor{black}
\multirow{4}{*}{2.90ha (50\%)} & Scratch & \cellcolor{rankgreen!24}\textcolor{black!65}{\textbf{57.9}} & \cellcolor{rankgreen!16}\textcolor{black!65}{79.0} & \cellcolor{rankgreen!24}\textcolor{black!65}{\textbf{74.2}} \\
 & Head Probing & \cellcolor{rankgreen!0}33.3 & \cellcolor{rankgreen!0}56.7 & \cellcolor{rankgreen!0}50.6 \\
 & Decoder Probing & \cellcolor{rankgreen!8}51.8 & \cellcolor{rankgreen!8}73.9 & \cellcolor{rankgreen!8}68.2 \\
 & Fine-tuning & \cellcolor{rankgreen!16}56.8 & \cellcolor{rankgreen!24}\textbf{80.8} & \cellcolor{rankgreen!16}74.0 \\

\arrayrulecolor{black!10}\midrule\arrayrulecolor{black}
\multirow{4}{*}{5.71ha (100\%)} & Scratch & \cellcolor{rankgreen!24}\textcolor{black!65}{\textbf{61.3}} & \cellcolor{rankgreen!24}\textcolor{black!65}{\textbf{82.0}} & \cellcolor{rankgreen!24}\textcolor{black!65}{\textbf{78.4}} \\
 & Head Probing & \cellcolor{rankgreen!0}38.7 & \cellcolor{rankgreen!0}60.9 & \cellcolor{rankgreen!0}54.8 \\
 & Decoder Probing & \cellcolor{rankgreen!8}55.2 & \cellcolor{rankgreen!8}76.9 & \cellcolor{rankgreen!8}71.2 \\
 & Fine-tuning & \cellcolor{rankgreen!16}60.6 & \cellcolor{rankgreen!16}81.4 & \cellcolor{rankgreen!16}77.6 \\

\bottomrule
\end{tabular}
}
\caption{
\textbf{Forest instance segmentation results on FOR-instanceV2 test set.} Comparison of training from scratch and transfer from the self-supervised pretrained backbone. The data percentage is computed based on the number of training samples and is not directly comparable to that in~\Cref{tab:sem_seg}, due to differences in the chunking strategy. Values in parentheses (shown once in the top block) denote each setup's trainable / total parameters, which are identical across data fractions.
}
\label{tab:ins_seg}
\end{table}

\noindent\textbf{Evaluation metrics.} 
Previous studies on forest point cloud instance segmentation~\citep{xiang2024_forainet,wielgosz2024_segmentanytree,henrich2024treelearn,xiang2025forestformer3d,nguyen2026forestmamba,wielgosz2026segmentanytreev2} commonly evaluate performance using the F1 score. A predicted instance is counted as a true positive (TP) if its best-matching ground-truth instance yields IoU $\geq 0.5$ and that ground-truth instance has not already been matched by a higher-confidence prediction; otherwise it is a false positive (FP). False negatives (FN) are ground-truth instances that remain unmatched. Precision and recall are then defined as:
\begin{align}
\mathrm{Precision} &= \frac{\mathrm{TP}}{\mathrm{TP}+\mathrm{FP}}\;,
\label{eq_precision}\\[2pt]
\mathrm{Recall} &= \frac{\mathrm{TP}}{\mathrm{TP}+\mathrm{FN}}\;,
\label{eq_recall}
\end{align}
and the F1 score is their harmonic mean:
\begin{equation}
\mathrm{F1} = \frac{2 \times \mathrm{Precision} \times \mathrm{Recall}}
{\mathrm{Precision}+\mathrm{Recall}}\;.
\label{eq_f1}
\end{equation}
A second metric commonly reported is mean coverage (Cov), which measures how well each ground-truth instance is recovered. Given ground-truth instances $\mathcal{G}=\{g_1,\ldots,g_{N_g}\}$ and predictions $\mathcal{P}=\{p_1,\ldots,p_{N_p}\}$, each $g_i$ is assigned its best-overlapping prediction and the resulting IoU is averaged:
\begin{equation}
\mathrm{Cov} = \frac{1}{N_g} \sum_{i=1}^{N_g}
\max_{j} \mathrm{IoU}(g_i, p_j)\;,
\label{eq_mucov}
\end{equation}
computed per plot and averaged across plots. Since overlap enters directly rather than through a threshold, partially recovered trees still contribute proportionally.

However, neither protocol characterises segmentation quality completely. The F1 score reflects performance at a single IoU threshold and ignores mask quality at other levels of overlap, while Cov is recall-oriented: it is computed per ground-truth instance and therefore does not penalise spurious predictions. Moreover, neither metric makes use of the confidence scores that rank predictions.
In the computer vision literature, it is more established to report average precision (AP) metrics, which evaluate performance over multiple recall levels and IoU thresholds. This provides a more comprehensive, application-agnostic assessment of segmentation quality. Here, we adopt AP-based metrics as the primary evaluation criteria. Predicted instances are ranked by their confidence scores in descending order. Cumulative TP and FP counts are computed along this ranked list, yielding a precision–recall curve. AP is the area under this curve approximated by 101-point interpolation following the COCO evaluation protocol~\citep{lin2014microsoft}:
\begin{equation}
\mathrm{AP} = \frac{1}{101} \sum_{r \in \{0,\, 0.01,\, \ldots,\, 1.0\}} \max_{\tilde{r} \geq r} P(\tilde{r})\;,
\label{eq_ap}
\end{equation}
where $P(\tilde{r})$ denotes the measured precision at recall level $\tilde{r}$. AP@25 and AP@50 evaluate this curve at IoU matching thresholds of 0.25 and 0.50, respectively. AP@25 captures coarse detection ability, which is useful for assessing whether the model identifies the approximate spatial extent of individual trees. AP@50 enforces stricter spatial agreement and is more sensitive to precise delineation of tree boundaries. To provide a threshold-agnostic summary, we also report the COCO mean Average Precision (mAP), computed by averaging AP across ten IoU thresholds from 0.50 to 0.95 in steps of 0.05:
\begin{equation}
\mathrm{mAP} = \frac{1}{10} \sum_{t \in \{0.50,\, 0.55,\, \ldots,\, 0.95\}} \mathrm{AP}_{t}\;.
\label{eq_map}
\end{equation}
This metric rewards models that produce not only correctly detected but also precisely delineated tree instances, making it a comprehensive indicator of overall instance segmentation quality. When directly comparing with prior forest point cloud studies, we also report the F1 score for completeness.

\begin{figure}[!t]%
\centering%
\includegraphics[width=0.9\linewidth]{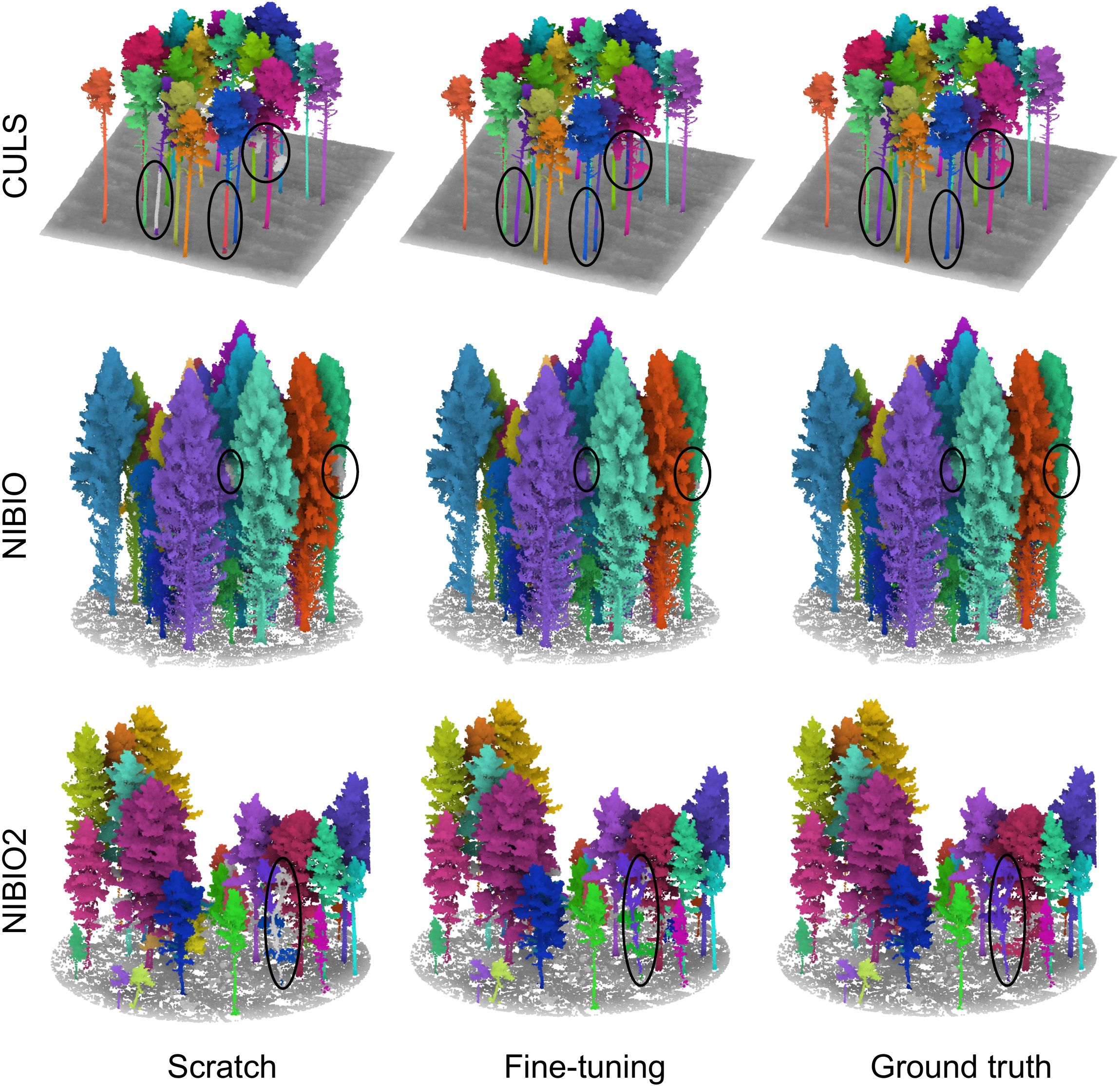}
\caption{\textbf{Qualitative forest instance segmentation results on FOR-instanceV2 test set samples when 1.35ha labelled training data is used.} Comparison of training from scratch and fine-tuning from the self-supervised pretrained backbone. Each ground-truth tree is given a unique color; predicted instances are matched to ground-truth via maximum-IoU Hungarian assignment and share the color of their matched ground-truth tree. Unmatched predictions (false positives) and unmatched ground-truth instances (missed trees) are shown in their own distinct colors. Non-tree (ground) points are colored light gray.}\label{fig:visuals_ins_seg}
\end{figure}

\noindent\textbf{Results.} \Cref{tab:ins_seg} reports the instance segmentation performance on the FOR-instanceV2 test set under varying proportions of labelled training data (5\%-100\%). In the low-data regime (0.29ha), fine-tuning provides a clear advantage over training from scratch, improving mAP from 26.6 to 28.2, highlighting the effectiveness of pretrained representations for instance-level understanding when annotations are limited. This observation is further supported by the qualitative results shown in \Cref{fig:visuals_ins_seg}, where the models are trained only using 1.35ha (20\%) labelled data. Both training from scratch and fine-tuning produce reasonable segmentation results, while fine-tuning yields more accurate instance predictions, correcting under-segmented trees, reducing incorrect instance assignments, and better preserving fine-grained details.

As the amount of labelled data increases, however, the performance gap between fine-tuning and training from scratch narrows. In higher-data regimes, training from scratch slightly outperforms fine-tuning. This suggests that, given sufficient labelled data, models trained from scratch can effectively learn instance-level representations without relying on pretraining. This behavior contrasts with the trends observed in our standalone semantic segmentation experiments in \Cref{subsec:sem}, where fine-tuning remains beneficial even with the full training set. Overall, these results suggest that the benefit of pretraining depends on both the downstream task and the amount of labelled data available.

Finally, we compare our approach with state-of-the-art methods in \Cref{tab:ins_seg_sota}. Since our instance segmentation framework jointly predicts instance and semantic segmentation, we report results for both tasks from a single model. On the full training set, LitePT-\emph{scratch} slightly outperforms ForPT-\emph{finetune}, while both variants surpass the previous best results in F1 score for individual tree segmentation and mIoU for semantic segmentation.

\begin{table}[!t]
\centering
\setlength{\tabcolsep}{2pt}
\resizebox{\columnwidth}{!}{%
\begin{tabular}{lccccccccc}
\toprule
& \multicolumn{5}{c}{Individual Tree Seg.} & \multicolumn{4}{c}{Semantic Seg.} \\
\cmidrule(r){2-6} \cmidrule(r){7-10}
Method  & mAP (\%) $\uparrow$ & mAP25 (\%) $\uparrow$ & mAP50 (\%) $\uparrow$ & F1 (\%) $\uparrow$ & Cov (\%) $\uparrow$ & Ground (\%) $\uparrow$ & Wood (\%) $\uparrow$ & Leaf (\%) $\uparrow$ & mIoU (\%) $\uparrow$ \\
\midrule

ForAINetV2\_R8~\citep{xiang2024_forainet} & 34.5 & 65.1 & 55.2 & 72.4 & 73.3 & 98.3 & 68.1 & 94.3 & 86.9 \\ %
ForAINetV2\_R16~\citep{xiang2024_forainet} & 37.3 & 63.8 & 54.8 & 70.8  & 72.7 & \textbf{98.8} & 69.2 & 94.5 & 87.5 \\ %
TreeLearn~\citep{henrich2024treelearn} & 19.6 & 41.4 & 30.9 & 50.3 & 52.5 & - & - & - & - \\ %
OneFormer3D~\citep{kolodiazhnyi2024oneformer3d} &  48.7 & 71.8 & 65.7 & 73.1 & 80.0 & 97.8 & 63.1 & 93.4 & 84.8 \\ %
ForestFormer3D~\citep{xiang2025forestformer3d} & 58.3 & 79.5 & 74.8 & 83.5 & 81.6 & 98.6 & 65.5 & 93.9 & 86.0 \\ %
ForestMamba~\citep{nguyen2026forestmamba} & 54.9 & 77.7 & 72.0 & 82.0 & 76.6 & 98.5 & 65.4 & 93.9 & 85.9\\ %
SegmentAnyTreeV2~\citep{wielgosz2026segmentanytreev2} & - & - & - & 83.4 & \textbf{89.6} & 99.1 & 70.5 & 93.9 & 87.8 \\ %
\arrayrulecolor{black!10}\midrule\arrayrulecolor{black}
LitePT-\emph{scratch} (ours) & \textbf{61.3} & \textbf{82.0} & \textbf{78.4} & \textbf{85.5} & 81.6 & 98.4 & \textbf{71.9} & \textbf{94.6} & \textbf{88.3} \\ %
ForPT-\emph{finetune} (ours) & 60.6 & 81.4 & 77.6 & 84.7  & 81.1 & 98.5 & 71.4 & 94.4 & 88.1 \\ %
\bottomrule
\end{tabular}%
}
\caption{
\textbf{Comparison with state-of-the-art methods on forest instance segmentation and semantic segmentation on FOR-instanceV2 test set.}
}
\label{tab:ins_seg_sota}
\end{table}

\subsection{Species classification}
\label{subsec:cls}

\begin{figure}[ht]%
\centering%
\includegraphics[width=\linewidth]{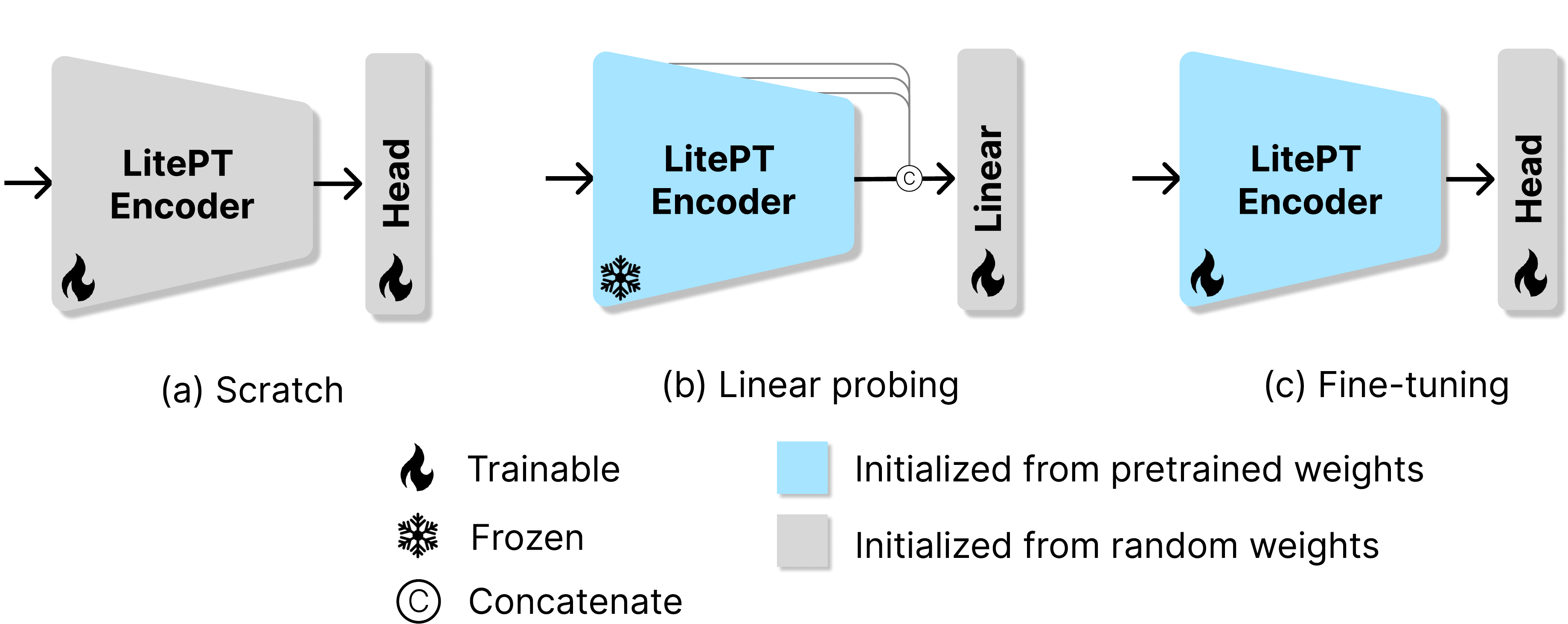}
\caption{\textbf{Model variants for single-tree species classification and age regression.} The task head is a lightweight MLP.}\label{fig:species_age_model_variants}
\end{figure}

\noindent\textbf{Setup.} In this task, the model takes a single-tree point cloud as input and predicts its species. For efficiency, each tree is subsampled to 8,192 points using farthest point sampling. 
Following the adaptation strategies shown in~\Cref{fig:species_age_model_variants}, we consider scratch training, linear probing, and fine-tuning.
We use a lightweight MLP as the classification head. 
In the \emph{scratch} setting, the classification head is attached to the encoder output, and all network weights are randomly initialized. 
In the \emph{linear probing} setting, the pretrained encoder is frozen, and global pooling is applied to the concatenated multi-scale encoder features, followed by a linear layer for classification. 
In the \emph{fine-tuning} setting, we initialize the encoder of the backbone with pretrained weights and train the model end-to-end. For the \emph{fine-tuning} setting in this task, we again observe that partially loading pretrained weights is beneficial; specifically, initializing the first three encoder stages yields the best performance. 

Additionally, we report results obtained using a non-parametric $k$NN probe. We run the frozen encoder, concatenate the multi-scale features and apply global average pooling to build a feature bank from all labelled training trees. Each test tree is classified by a majority vote over its $k$ nearest bank entries under cosine similarity. We use $k=1$ for all data fractions, as larger $k$ is biased towards frequent species in the long-tailed dataset.

\begin{table}[!t]
\centering
\setlength{\tabcolsep}{35pt}
\resizebox{\columnwidth}{!}{
\begin{tabular}{clcc}
\toprule
\cmidrule{1-4} 
& & \multicolumn{2}{c}{Overall}  \\
 \cmidrule(r){3-4}
Training data & Setup~{\scriptsize(trainable/total parameters)}   & mAcc $\uparrow$ & allAcc $\uparrow$ \\
\midrule
\multirow{4}{*}{141 trees (1\%)} & Scratch~{\scriptsize(12.6M/12.6M)} & \cellcolor{rankgreen!0}\textcolor{black!65}{22.8} & \cellcolor{rankgreen!8}\textcolor{black!65}{46.2} \\
 & $k$NN~{\scriptsize(0/12.4M)} & \cellcolor{rankgreen!8}25.9 & \cellcolor{rankgreen!0}38.6 \\
 & Linear Probing~{\scriptsize(33.3K/12.4M)} & \cellcolor{rankgreen!24}\textbf{32.8} & \cellcolor{rankgreen!16}50.1 \\
 & Fine-tuning~{\scriptsize(12.6M/12.6M)} & \cellcolor{rankgreen!16}29.2 & \cellcolor{rankgreen!24}\textbf{50.8} \\

\arrayrulecolor{black!10}\midrule\arrayrulecolor{black}
\multirow{4}{*}{708 trees (5\%)} & Scratch & \cellcolor{rankgreen!16}\textcolor{black!65}{45.7} & \cellcolor{rankgreen!16}\textcolor{black!65}{66.4} \\
 & $k$NN & \cellcolor{rankgreen!0}34.8 & \cellcolor{rankgreen!0}51.5 \\
 & Linear Probing & \cellcolor{rankgreen!8}42.4 & \cellcolor{rankgreen!8}62.0 \\
 & Fine-tuning & \cellcolor{rankgreen!24}\textbf{48.7} & \cellcolor{rankgreen!24}\textbf{67.5} \\

\arrayrulecolor{black!10}\midrule\arrayrulecolor{black}
\multirow{4}{*}{1416 trees (10\%)} & Scratch & \cellcolor{rankgreen!16}\textcolor{black!65}{57.3} & \cellcolor{rankgreen!24}\textcolor{black!65}{\textbf{73.2}} \\
 & $k$NN & \cellcolor{rankgreen!0}38.5 & \cellcolor{rankgreen!0}55.6 \\
 & Linear Probing & \cellcolor{rankgreen!8}47.9 & \cellcolor{rankgreen!8}65.0 \\
 & Fine-tuning & \cellcolor{rankgreen!24}\textbf{57.9} & \cellcolor{rankgreen!16}73.0 \\

\arrayrulecolor{black!10}\midrule\arrayrulecolor{black}
\multirow{4}{*}{2833 trees (20\%)} & Scratch & \cellcolor{rankgreen!16}\textcolor{black!65}{62.4} & \cellcolor{rankgreen!16}\textcolor{black!65}{77.7} \\
 & $k$NN & \cellcolor{rankgreen!0}43.6 & \cellcolor{rankgreen!0}58.6 \\
 & Linear Probing & \cellcolor{rankgreen!8}53.6 & \cellcolor{rankgreen!8}67.4 \\
 & Fine-tuning & \cellcolor{rankgreen!24}\textbf{64.8} & \cellcolor{rankgreen!24}\textbf{77.8} \\

\arrayrulecolor{black!10}\midrule\arrayrulecolor{black}
\multirow{4}{*}{7082 trees (50\%)} & Scratch & \cellcolor{rankgreen!16}\textcolor{black!65}{73.3} & \cellcolor{rankgreen!16}\textcolor{black!65}{83.1} \\
 & $k$NN & \cellcolor{rankgreen!0}49.4 & \cellcolor{rankgreen!0}63.2 \\
 & Linear Probing & \cellcolor{rankgreen!8}60.2 & \cellcolor{rankgreen!8}71.1 \\
 & Fine-tuning & \cellcolor{rankgreen!24}\textbf{74.7} & \cellcolor{rankgreen!24}\textbf{83.3} \\

\arrayrulecolor{black!10}\midrule\arrayrulecolor{black}
\multirow{4}{*}{14165 trees (100\%)} & Scratch & \cellcolor{rankgreen!16}\textcolor{black!65}{79.9} & \cellcolor{rankgreen!16}\textcolor{black!65}{86.1} \\
 & $k$NN & \cellcolor{rankgreen!0}53.8 & \cellcolor{rankgreen!0}66.0 \\
 & Linear Probing & \cellcolor{rankgreen!8}63.1 & \cellcolor{rankgreen!8}72.8 \\
 & Fine-tuning & \cellcolor{rankgreen!24}\textbf{80.1} & \cellcolor{rankgreen!24}\textbf{86.4} \\

\bottomrule
\end{tabular}
}
\caption{
\textbf{Tree species classification results on FOR-species20K validation set.} Comparison of training from scratch and transfer from the self-supervised pretrained backbone. Values in parentheses (shown once in the top block) denote each setup's trainable / total parameters, which are identical across data fractions.
}
\label{tab:species_cls}
\end{table}

\noindent\textbf{Evaluation metrics.} We report mean class accuracy (mAcc) and overall accuracy (allAcc), following the same definitions as in the semantic segmentation evaluation, as defined in \Cref{eq_macc,all_macc}. mAcc evaluates the averaged classification performance across all species and therefore is less sensitive to class imbalance and long-tail species distributions. In contrast, allAcc measures the overall classification accuracy across the entire evaluation set and is more influenced by dominant species classes with larger numbers of samples. Reporting both metrics provides a comprehensive evaluation of classification performance for both balanced per-class prediction and overall dataset-level accuracy.

\noindent\textbf{Results.} 
\Cref{tab:species_cls} presents the tree species classification performance on the FOR-species20K validation set under varying proportions of labelled training data (1\%-100\%). In the extremely low-data regime (only 141 trees, 1\%), both fine-tuning and linear probing outperform training from scratch,  demonstrating the clear benefit of pretrained representations when labelled data are scarce. As the amount of labelled data increases, the performance gap between fine-tuning and training from scratch gradually diminishes. Nevertheless, pretraining continues to provide important advantages in terms of optimization efficiency. Specifically, \Cref{fig:val_curve} (b) shows the validation overall accuracy over training epochs for the three training strategies when 100\% data is used. Both fine-tuning and linear probing achieve strong performance early in training, exhibiting significantly faster convergence and more stable optimization compared with training from scratch. These results demonstrate that even when the final performance differences between fine-tuning and training from scratch are small, pretraining substantially improves training efficiency by accelerating convergence and stabilizing optimization, making it particularly advantageous in resource-constrained or time-sensitive scenarios.

\begin{figure}[ht]%
\centering%
\includegraphics[width=\linewidth]{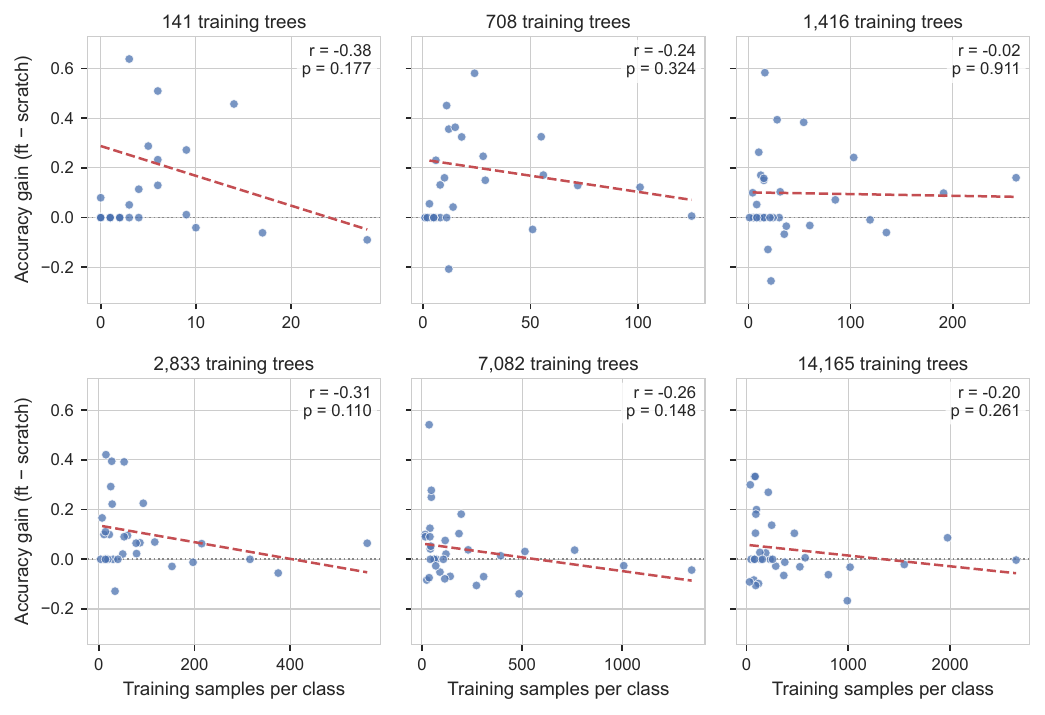}
\caption{\textbf{Per-class accuracy gain of fine-tuning the self-supervised pretrained backbone over training from scratch at epoch 20}, plotted against the number of training samples per class for six data regimes. Each point represents one tree species. Dashed lines show linear regression fits (r and p-value annotated); classes where both models scored zero are excluded from regression.}\label{fig:species_gain_vs_sample}
\end{figure}

As FOR-species20K exhibits a pronounced long-tail distribution, we further analyze the per-class accuracy gain of fine-tuning over training from scratch. \Cref{fig:species_gain_vs_sample} plots the accuracy improvement (fine-tuning minus scratch) against the number of training samples per class at epoch 20 across different training data regime (1\%-100\%). Overall, we observe a negative correlation between accuracy gain and the number of training samples per class, particularly in low-data regimes. The fitted correlation is negative in every one of the six label regimes ($r$ between $-0.02$ and $-0.38$), although no individual regime reaches statistical significance (all $p>0.10$). It is therefore the consistency of the sign across six independent label regimes, rather than any single fit, that suggests fine-tuning provides larger improvements for underrepresented classes, where learning from scratch is more challenging due to limited supervision. These results suggest that pretraining improves overall performance in part by mitigating the long-tail effect, disproportionately benefiting classes with fewer training samples. This property is practically valuable in real-world ecological datasets, where class imbalance is common and rare species are of high importance.

We further compare our approach with previous state-of-the-art methods on the FOR-species20K dataset. For benchmarking purposes, we make the following modifications to the setup we used in the data-efficiency experiments. First, instead of subsampling each tree to 8,192 points, we use full point cloud as input, which improves final performance. Second, we remove the slight random rotation augmentation around $x$ and $y$ axes, as it incurs significant computational overhead when processing full point clouds. Third, we utilize all officially provided development data from FOR-species20K for training. 
As shown in \Cref{tab:species_cls_leadboard}, LitePT-\emph{scratch} and ForPT-\emph{finetune} outperform all previous methods on this benchmark. LitePT-\emph{scratch} achieves an overall accuracy of 82.7\%, while fine-tuning the pretrained ForPT backbone further increases the accuracy to 83.2\%, achieving the best performance across all reported metrics. In the frozen-feature setting, ForPT-\emph{linear probing} achieves an overall accuracy of 68.7\%, demonstrating that the learned representations retain useful information for tree species classification even without fine-tuning.

\begin{table}[ht]
\centering
\setlength{\tabcolsep}{5pt}
\resizebox{\columnwidth}{!}{
\begin{tabular}{lccccc}
\toprule
Method & Input data & Overall Accuracy (\%) $\uparrow$ & Precision (\%) $\uparrow$ & Recall (\%) $\uparrow$ & F1 (\%) $\uparrow$ \\
\midrule
SimpleView & Image & 76.2 & 76.9 & 75.5 & 75.6  \\
YOLOv5 & Image & 77.9 & 84.2 & 75.0 & 77.3 \\
DetailView & Image & 79.5 & 82.3 & 76.7 & 78.0 \\
\midrule
PointAugment + DGCNN & Point cloud & 68.3 & 72.5 & 65.7 & 70.3 \\
PointMixer & Point cloud & 71.1 & 74.4 & 65.5 & 71.1  \\
MinkNet & Point cloud & 73.7 & 79.9 & 70.6 & 72.3 \\
Ensemble PointNet++ & Point cloud & 75.6 & 78.2 & 73.5 & 74.9  \\
\midrule
LitePT-\emph{scratch} (ours) & Point cloud & 82.7 & 85.3 & 81.6 & 82.5 \\
ForPT-\emph{linear probing} (ours) & Point cloud & 68.7 & 74.2 & 66.0 & 67.8 \\
ForPT-\emph{finetune} (ours) & Point cloud & \textbf{83.2} & \textbf{86.0} & \textbf{81.8} & \textbf{83.0} \\

\bottomrule
\end{tabular}
}
\caption{
\textbf{Comparison with state-of-the-art methods on tree species classification on FOR-species20K test set.} Scores of prior work courtesy of ~\cite{puliti2025_forspecies20k}.  
}
\label{tab:species_cls_leadboard}
\end{table}

\subsection{Age regression}
\label{subsec:age}

\noindent\textbf{Setup.} In this task, the model takes a single-tree point cloud as input and predicts its age. As in the species classification task, each tree is subsampled to 8,192 points using farthest point sampling before being fed into the model.
The evaluated training strategies are shown in~\Cref{fig:species_age_model_variants}.
We use a lightweight MLP as the regression head. 
In the \emph{scratch} setting, the regression head is attached to the final layer of the encoder and all network weights are randomly initialized. 
In the \emph{linear probing} setting, we freeze the pretrained encoder and apply global pooling to the concatenated multi-scale encoder features, which is then fed into a single linear layer for regression. 
In the \emph{fine-tuning} setting, we initialize the encoder of the backbone with pretrained weights and train the model end-to-end. Same with the fine-tuning strategy with species classification, we only initialize the first three encoder stages.

\noindent\textbf{Evaluation metrics.}
Consistent with previous work on tree age regression~\citep{PULITI2026115462}, we report the root mean squared error (RMSE), the mean difference (MD), and the coefficient of determination ($\mathrm{R}^2$). RMSE quantifies the overall prediction error by measuring the average magnitude of the residuals, MD evaluates the systematic errors in the predictions, and $\mathrm{R}^2$ measures the proportion of variance in the reference ages explained by the model predictions. These metrics are defined as follows:
\begin{equation}
\mathrm{RMSE} = \sqrt{\frac{\sum_{i=1}^{N} (\hat{y}_i - y_i)^2}{N}}\;,
\label{eq_rmse}
\end{equation}
\begin{equation}
\mathrm{MD} = \frac{\sum_{i=1}^{N} \hat{y}_i - y_i}{N}\;,
\label{eq_md}
\end{equation}
\begin{equation}
\mathrm{R}^2 = 1 -
\frac{\sum_{i=1}^{N} (y_i - \hat{y}_i)^2}
{\sum_{i=1}^{N} (y_i - \bar{y})^2}\;,
\label{eq_r2}
\end{equation}

\noindent where $\hat{y}_i$ and $y_i$ are the predicted and reference age for the
$i^{\text{th}}$ tree, and $N$ is the total number of trees in the evaluation data. Lower RMSE values indicate improved predictive accuracy, MD values closer to zero indicate reduced systematic bias, and higher $\mathrm{R}^2$ values indicate a better fit between predicted and reference ages.

\begin{table}[ht]
\centering
\setlength{\tabcolsep}{20pt}
\resizebox{\columnwidth}{!}{
\begin{tabular}{clccc}
\toprule
Training data & Setup~{\scriptsize(trainable/total parameters)}  &  RMSE (years) $\downarrow$ & MD (years) $\rightarrow 0$ & $\mathrm{R}^{2}$ $\uparrow$ \\
\midrule
\multirow{3}{*}{62 trees (5\%)} & Scratch~{\scriptsize(12.6M/12.6M)} & \cellcolor{rankgreen!12}\textcolor{black!65}{47.07} & \cellcolor{rankgreen!12}\textcolor{black!65}{-30.48} & \cellcolor{rankgreen!12}\textcolor{black!65}{-0.20} \\
 & Linear Probing~{\scriptsize(1K/12.4M)} & \cellcolor{rankgreen!24}\textbf{32.40} & \cellcolor{rankgreen!24}\textbf{-5.87} & \cellcolor{rankgreen!24}\textbf{0.43} \\
 & Fine-tuning~{\scriptsize(12.6M/12.6M)} & \cellcolor{rankgreen!0}48.23 & \cellcolor{rankgreen!0}-31.99 & \cellcolor{rankgreen!0}-0.26 \\

\arrayrulecolor{black!10}\midrule\arrayrulecolor{black}
\multirow{3}{*}{125 trees (10\%)} & Scratch & \cellcolor{rankgreen!0}\textcolor{black!65}{33.87} & \cellcolor{rankgreen!0}\textcolor{black!65}{-18.03} & \cellcolor{rankgreen!0}\textcolor{black!65}{0.38} \\
 & Linear Probing & \cellcolor{rankgreen!24}\textbf{31.25} & \cellcolor{rankgreen!24}\textbf{-4.22} & \cellcolor{rankgreen!24}\textbf{0.47} \\
 & Fine-tuning & \cellcolor{rankgreen!12}32.96 & \cellcolor{rankgreen!12}-16.57 & \cellcolor{rankgreen!12}0.41 \\

\arrayrulecolor{black!10}\midrule\arrayrulecolor{black}
\multirow{3}{*}{250 trees (20\%)} & Scratch & \cellcolor{rankgreen!12}\textcolor{black!65}{27.10} & \cellcolor{rankgreen!12}\textcolor{black!65}{-5.54} & \cellcolor{rankgreen!12}\textcolor{black!65}{0.60} \\
 & Linear Probing & \cellcolor{rankgreen!0}30.77 & \cellcolor{rankgreen!24}\textbf{-5.27} & \cellcolor{rankgreen!0}0.49 \\
 & Fine-tuning & \cellcolor{rankgreen!24}\textbf{26.60} & \cellcolor{rankgreen!0}-8.68 & \cellcolor{rankgreen!24}\textbf{0.62} \\

\arrayrulecolor{black!10}\midrule\arrayrulecolor{black}
\multirow{3}{*}{625 trees (50\%)} & Scratch & \cellcolor{rankgreen!12}\textcolor{black!65}{24.09} & \cellcolor{rankgreen!12}\textcolor{black!65}{2.56} & \cellcolor{rankgreen!12}\textcolor{black!65}{0.69} \\
 & Linear Probing & \cellcolor{rankgreen!0}29.73 & \cellcolor{rankgreen!0}-2.95 & \cellcolor{rankgreen!0}0.52 \\
 & Fine-tuning & \cellcolor{rankgreen!24}\textbf{21.61} & \cellcolor{rankgreen!24}\textbf{1.78} & \cellcolor{rankgreen!24}\textbf{0.75} \\

\arrayrulecolor{black!10}\midrule\arrayrulecolor{black}
\multirow{3}{*}{1250 trees (100\%)} & Scratch & \cellcolor{rankgreen!24}\textcolor{black!65}{\textbf{19.27}} & \cellcolor{rankgreen!24}\textcolor{black!65}{\textbf{0.07}} & \cellcolor{rankgreen!24}\textcolor{black!65}{\textbf{0.80}} \\
 & Linear Probing & \cellcolor{rankgreen!0}29.14 & \cellcolor{rankgreen!0}-3.29 & \cellcolor{rankgreen!0}0.54 \\
 & Fine-tuning & \cellcolor{rankgreen!12}19.72 & \cellcolor{rankgreen!12}-1.21 & \cellcolor{rankgreen!12}0.79 \\

\bottomrule
\end{tabular}
}
\caption{
\textbf{Tree age regression results on FOR-age validation set.} Comparison of training from scratch and transfer from the self-supervised pretrained backbone. Values in parentheses (shown once in the top block) denote each setup's trainable / total parameters, which are identical across data fractions.
}
\label{tab:age_regression}
\end{table}

\noindent\textbf{Results.} \Cref{tab:age_regression} reports tree age regression performance on the FOR-age validation set under different training data fractions. When only 62 trees (5\%) or 125 trees (10\%) are available for training, linear probing substantially outperforms both training from scratch and full fine-tuning. For example, with only 62 training trees (5\%), linear probing reduces the RMSE from 47.07 years (scratch) and 48.23 years (fine-tuning) to 32.40 years while with only 1K trainable parameters. 

\begin{figure}[!t]%
\centering%
\includegraphics[width=\linewidth]{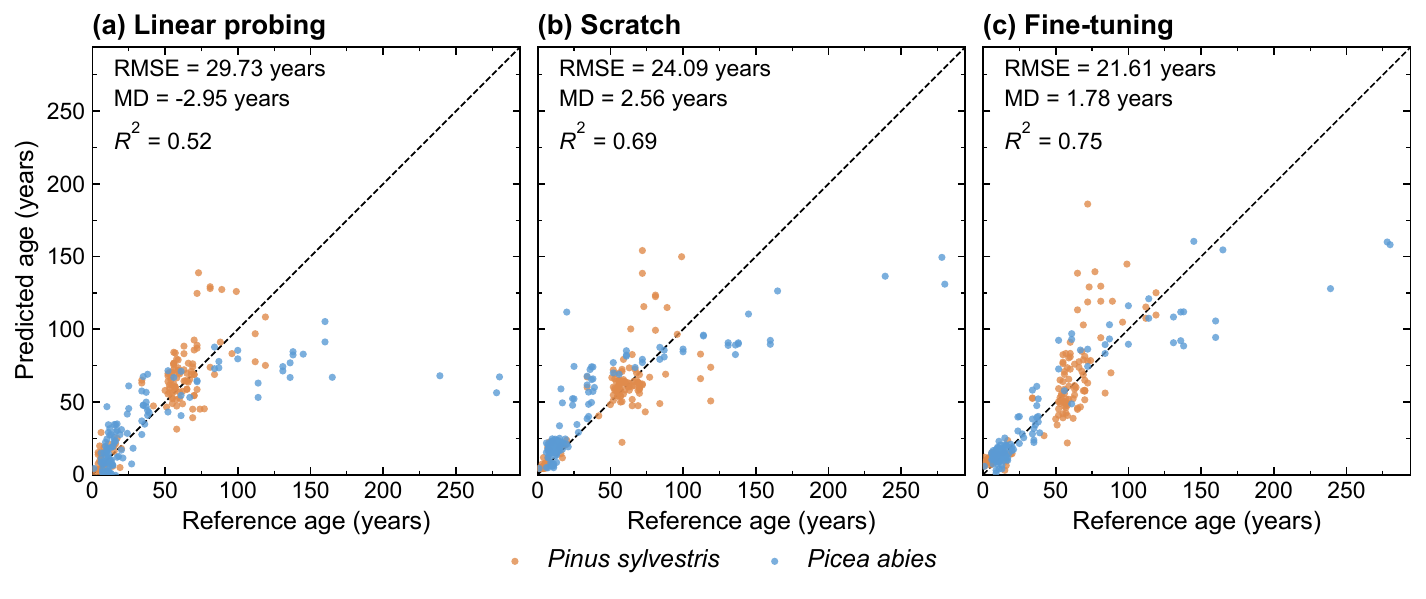}
\caption{\textbf{Scatterplots of the predicted vs reference age on the FOR-age validation set when 625 trees are used for training.} Comparison of training from scratch and transfer from the self-supervised pretrained backbone.}\label{fig:scatterplot_age}
\end{figure}

With larger training sets, full fine-tuning becomes increasingly effective and achieves the lowest RMSE in the 20\% and 50\% data regime. \Cref{fig:scatterplot_age} show the scatterplots of the predicted vs reference age when 625 trees (50\%) are used for training. 
Fine-tuning produces predictions that are most closely aligned with the 1:1 reference line, resulting in the lowest RMSE (21.61 years) and highest $R^2$ (0.75). In comparison, the model training from scratch exhibits a larger spread around the reference line, whereas linear probing shows greater underestimation of older trees. All the three plots show larger prediction errors for old trees, likely reflecting their underrepresentation in the long-tailed age distribution of the training data.

When the full training set (1250 trees) is used, training from scratch and fine-tuning achieve similar performance, both of which outperforms linear probing. Overall, linear probing is most effective in the low-data regime, whereas full fine-tuning becomes the preferred strategy as the amount of the labelled training data increases.

\begin{table}[!t]
\centering
\setlength{\tabcolsep}{25pt}
\resizebox{\columnwidth}{!}{
\begin{tabular}{lccc}
\toprule
Method & RMSE (years) $\downarrow$ & MD (years) $\rightarrow 0$ & $\mathrm{R}^{2}$ $\uparrow$ \\
\midrule
Linear regression &  34.15 & -2.68 & 0.28  \\
PTv3-\emph{scratch} &  22.80 & -3.96 & 0.68  \\
ForestFormer3D-\emph{finetune} &  20.71 & 2.37 & 0.74\\
\midrule
LitePT-\emph{scratch} (ours) & 20.41 & \textbf{-0.55} & 0.74 \\
ForPT-\emph{linear probing} (ours) & 30.65 & -1.10 & 0.42 \\
ForPT-\emph{finetune} (ours) & \textbf{20.03} & 1.39 & \textbf{0.75}\\

\bottomrule
\end{tabular}
}
\caption{
\textbf{Comparison with state-of-the-art methods on forest age regression on FOR-age test set.} Scores of prior work courtesy of~\citet{PULITI2026115462} 
}
\label{tab:age_sota}
\end{table}

We report the results on the FOR-age test set and compare them with previous state-of-the-art approaches in \Cref{tab:age_sota}. 
LitePT-\emph{scratch} achieves an RMSE of 20.41 years and an $R^2$ of 0.74, already outperforming the previously reported PTv3-\emph{scratch} and ForestFormer3D-\emph{finetune} results~\citep{PULITI2026115462}. Fine-tuning the pretrained ForPT backbone improves the performance further, reducing the RMSE to 20.03 years and increasing $R^2$ to 0.75, thereby achieving the best overall performance.

\begin{figure}[!t]%
\centering%
\includegraphics[width=0.7\linewidth]{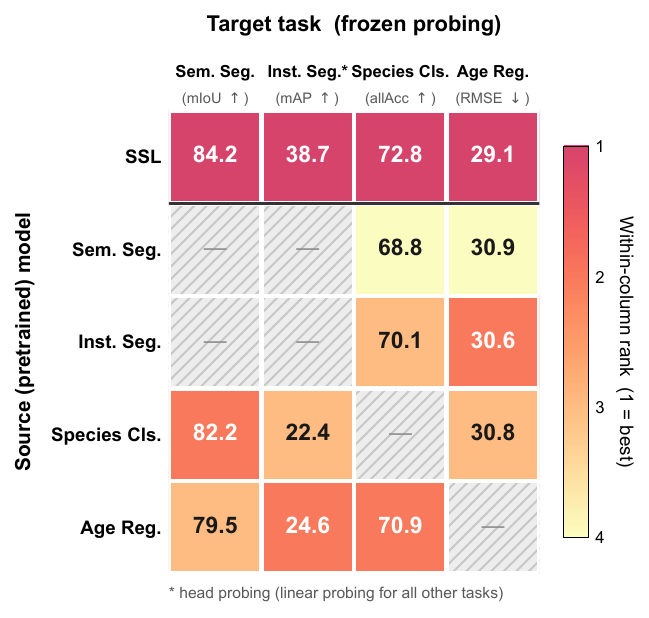}
\caption{\textbf{Cross-task transfer of frozen representations.} Each source model (row) is frozen-probed on four target tasks (column); color encodes within-column ranks (1=best) and printed number is the raw metric value. All probes are linear except head probing for instance segmentation. Because the ForestFormer3D head jointly trains a semantic-segmentation branch and an instance-segmentation branch, transfer between the two is not independent in either direction; we therefore omit the Inst.\ Seg.\ $\leftrightarrow$ Sem.\ Seg.\ transfer. }\label{fig:transfer_rank_matrix}
\end{figure}

\subsection{Cross-task transfer from supervised pretraining}
\label{subsec:supervised_transfer}

In the previous sections, we compared self-supervised pretraining against training from scratch for each downstream task. Here, we investigate the transferability of task-specific supervised pretraining. Specifically, we reuse the fully supervised models trained for the corresponding training-from-scratch baselines in \Cref{subsec:sem,subsec:ins,subsec:cls,subsec:age} as source-task pretrained models. After discarding the task-specific prediction module, the pretrained encoder is transferred to the remaining unseen downstream tasks through frozen probing. We compare these representations against self-supervised pretrained backbones. To isolate representation quality with minimal task-specific adaptation, we adopt linear probing for forest semantic segmentation, tree species classification, tree age regression, and head probing for forest instance segmentation. 

\Cref{fig:transfer_rank_matrix} summarizes the overall cross-task transferability. Across all four downstream tasks, self-supervised pretraining consistently produces the most transferable representations, outperforming task-specific supervised pretraining under the same probing protocol. In contrast, supervised pretraining remains strongly task-dependent: no single supervised source task transfers consistently well across all unseen downstream tasks. These results suggest that self-supervised pretraining is a more promising strategy for learning generic representations and moving toward foundation models for forest point clouds.

\begin{figure}[!t]%
\centering%
\includegraphics[width=\linewidth]{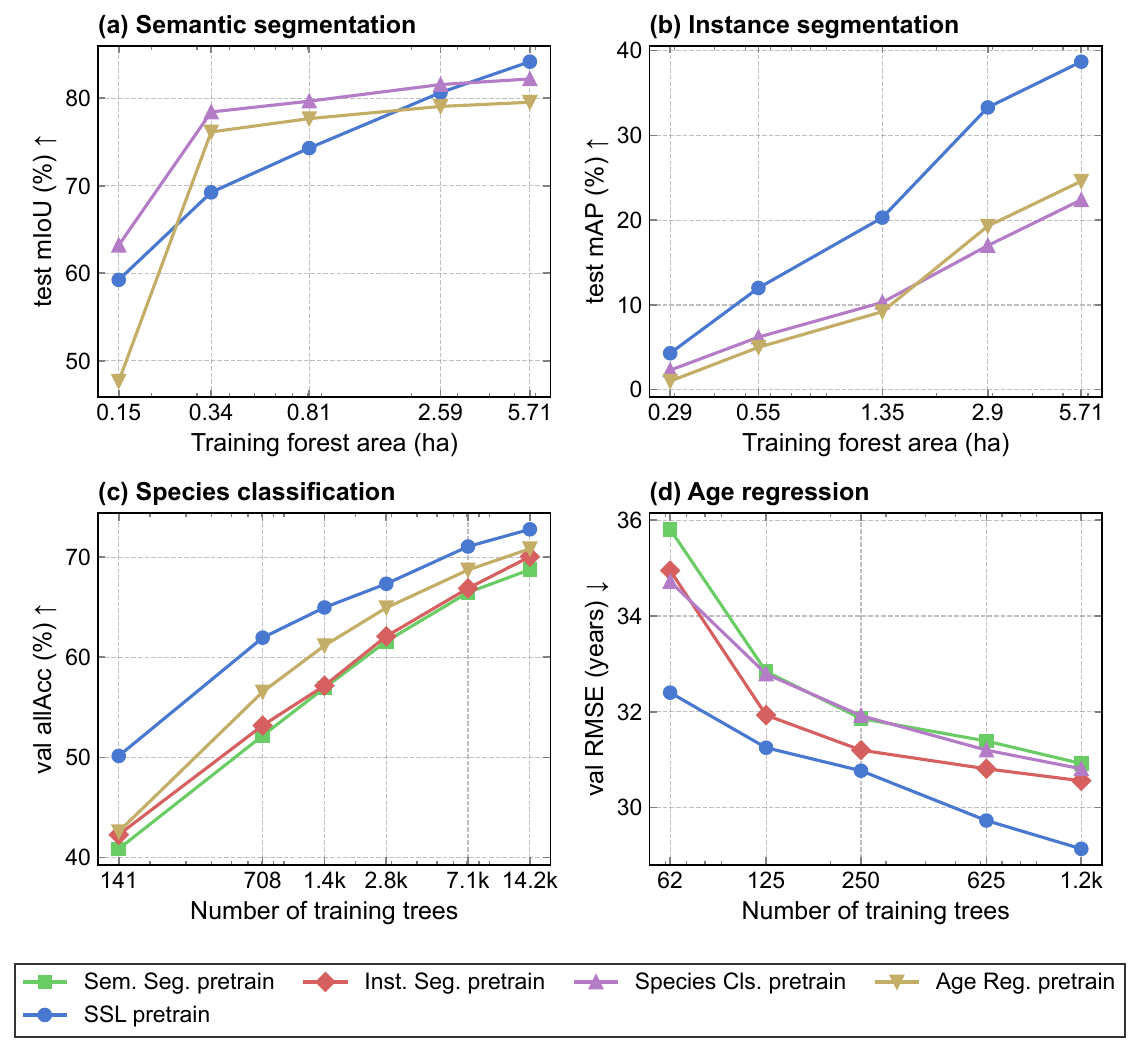}
\caption{\textbf{Cross-task transfer of frozen representations under varying annotation budgets.} All probes for pretrained representations are linear except head probing for instance segmentation.  }\label{fig:data_efficiency_all}
\end{figure}

\Cref{fig:data_efficiency_all} further analyzes transferability under different annotation budgets. Interestingly, models pretrained on tree species classification or tree age regression achieves higher linear probing performance on forest semantic segmentation than self-supervised pretraining in low-data regimes. 
This suggests that supervision from species classification and age regression encourages representations that capture semantically meaningful tree characteristics useful for semantic segmentation.
Nevertheless, this benefit does not generalize consistently to the other downstream tasks.

\section{Discussion}
\label{sec_discussion}

\subsection{Analysis and ablation studies}
\label{subsec:analysis}

\begin{figure}[ht]%
\centering%
\includegraphics[width=\linewidth]{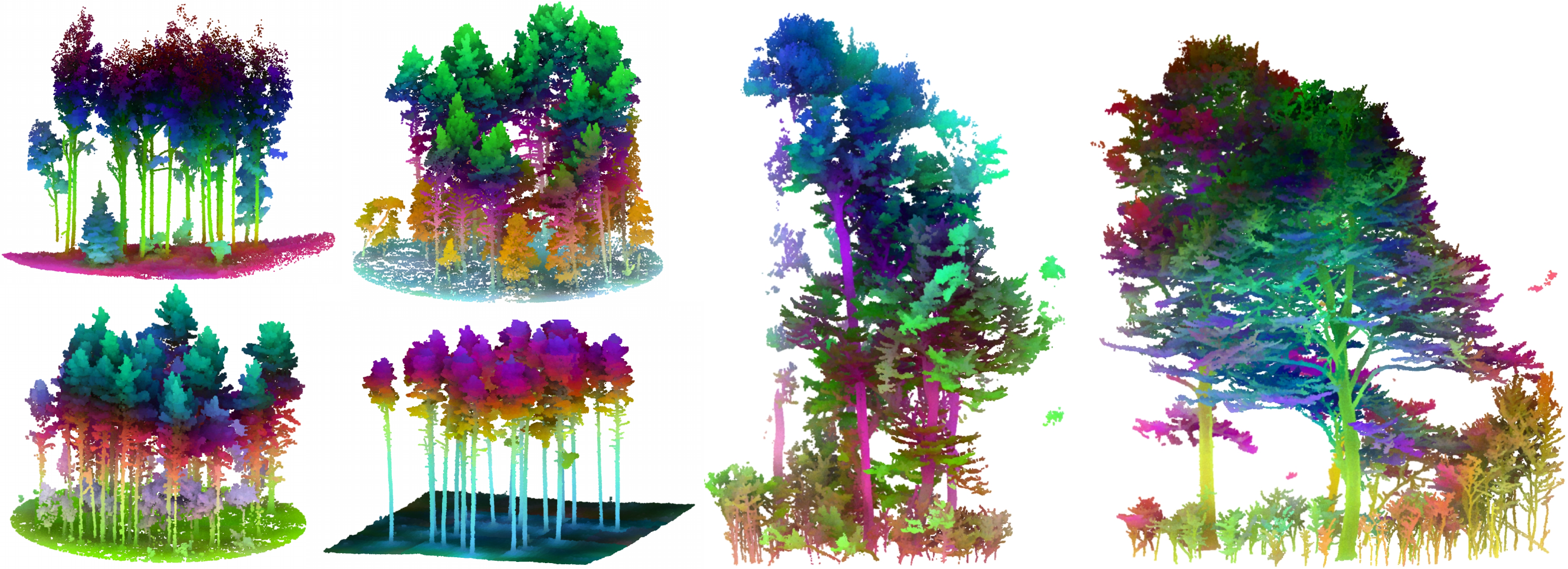}
\caption{\textbf{Learned representations through self-supervised pretraining.} Concatenated multi-scale encoder features are projected into a three-dimensional space using PCA, which is used as per-point color for rendering.}\label{fig:pca_features}
\end{figure}

\noindent\textbf{Learned representations.} We visualize the learned representations through self-supervised pretraining in \Cref{fig:pca_features}. Specifically, we concatenate multi-scale encoder features and project them into a three-dimensional space using principal component analysis (PCA), which is then used as per-point color for rendering.

As shown in the PCA visualizations, different semantic components are clearly distinguishable: leaves, wood, and ground exhibit distinct feature (color) values, indicating that the model learns separable representations for each class. Importantly, these features are not solely driven by height. For example, in scenes where trees vary significantly in height, the learned representations remain more consistent with vegetation components than with their vertical position. This suggests that self-supervised pretraining captures semantically meaningful features, which also offers an explanation why a simple $k$NN or linear probe already achieves largely correct semantic segmentation.

\noindent

\begin{figure}[!t]%
\centering%
\includegraphics[width=\linewidth]{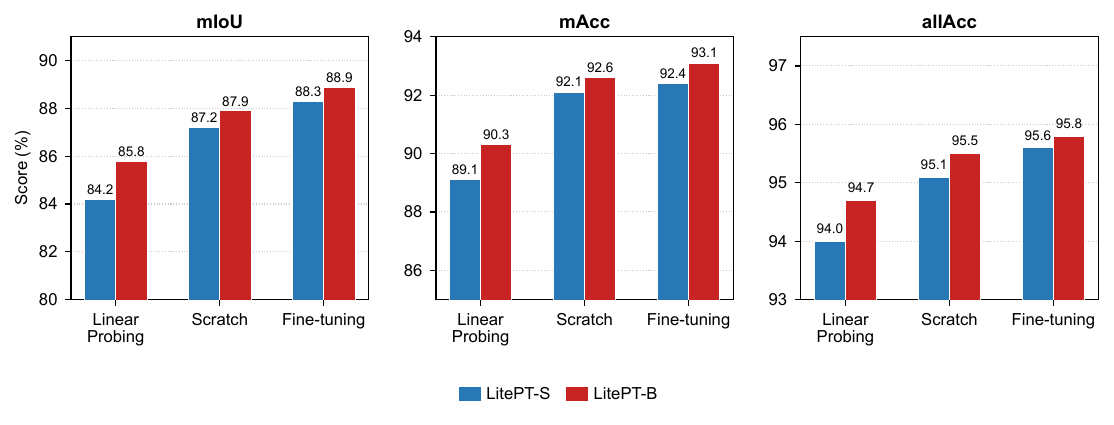}
\caption{\textbf{Model scaling on forest semantic segmentation.} We compare the pretrained LitePT-S encoder (12.4M) with LitePT-B encoder (44.4M). Comparison of training from scratch and transfer from the self-supervised pretrained backbone.}\label{fig:sem_seg_scaling_model}
\end{figure}

\noindent\textbf{Model scaling.}
In the main experiments, we pretrain LitePT-S (12.4M parameters). To study the effect of model scaling, we further pretrain a larger variant, LitePT-B (44.4M parameters). The results on forest semantic segmentation are shown in \Cref{fig:sem_seg_scaling_model}. Overall, scaling the model consistently improves performance across all training strategies and evaluation metrics. The improvements are especially notable in the linear probing setting, where mIoU increases from 84.2 to 85.8, indicating that larger pretrained models encode more expressive and transferable features. While scaling also leads to improved performance in both scratch and fine-tuning settings, the gains are relatively modest compared to the increase in model size. Therefore, considering the trade-off between performance and efficiency, we recommend LitePT-S as a practical choice for most applications.

\begin{figure}[!t]%
\centering%
\includegraphics[width=\linewidth]{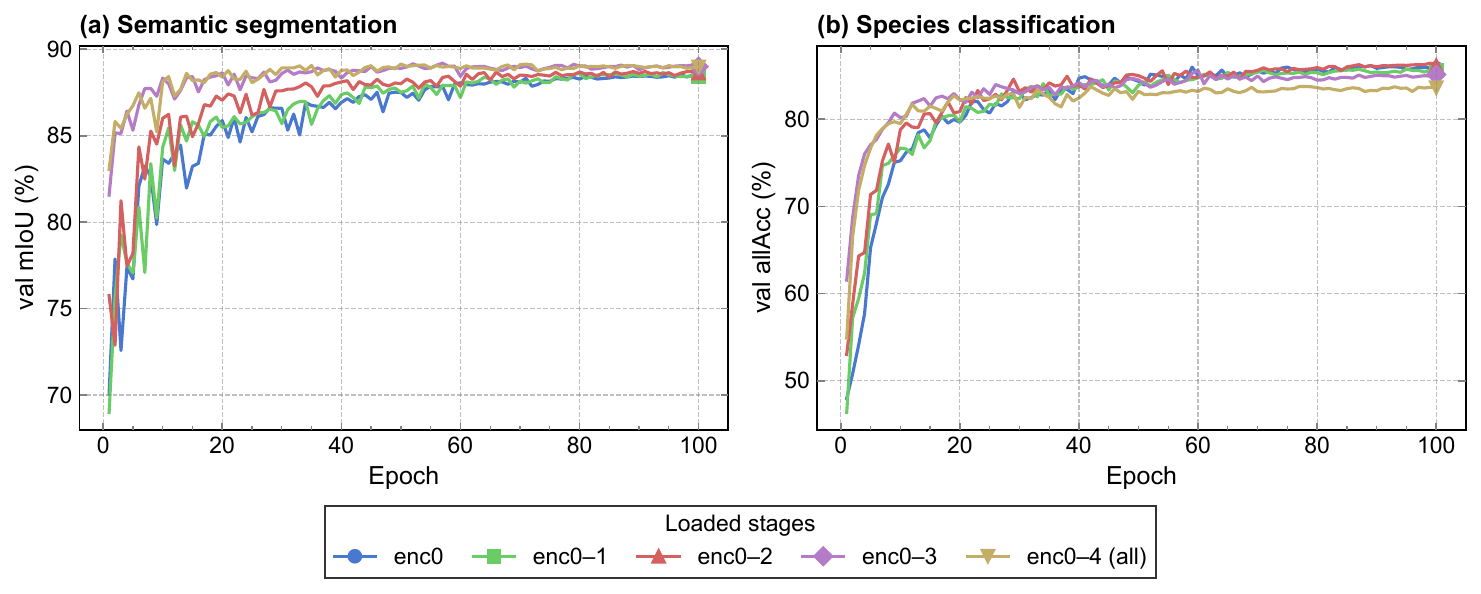}
\caption{\textbf{Validation performance over training epochs when partial loading different stages of the self-supervised pretrained encoder}, on (a) semantic segmentation (mIoU, \%); (b) tree species classification (overall accuracy, \%).}
\label{fig:val_curve_load_stage}
\end{figure}

\noindent\textbf{Partial loading of pretrained weights.} 
For certain downstream tasks (e.g., forest instance segmentation and tree species classification), we observe that, under the \emph{fine-tuning} setup, loading only a subset of pretrained weights (rather than the entire model) improves performance. 
In \Cref{fig:val_curve_load_stage}, we analyze the impact of partially loading pretrained encoder stages during downstream transfer learning. LitePT-S~\citep{yuelitept2026} consists of five encoding stages, and we progressively load pretrained weights from shallow to deeper stages (e.g., enc0-1 indicates loading stages 0 and 1). As shown in \Cref{fig:val_curve_load_stage} (a), for semantic segmentation, loading more stages generally improves performance and accelerates convergence. In contrast, for species classification as shown in \Cref{fig:val_curve_load_stage} (b), loading all pretrained stages leads to faster convergence but a lower final performance, whereas loading only early stages results in better final accuracy at the cost of slower convergence. Empirically, loading pretrained weights up to the third stage (i.e., enc0-2) provides the best trade-off between convergence speed and final performance.

We hypothesize that the hierarchical encoder captures representations at different levels of abstraction: early stages primarily encode local geometric structures and fine-grained details, whereas deeper stages focus on more global and shared semantic features. For semantic segmentation, these high-level semantic features are particularly beneficial, whereas for instance-discriminative tasks such as species classification, local geometric cues are more informative, and high-level shared semantics may be less relevant or even detrimental to adaptation.

Based on these findings, we recommend selectively loading pretrained weights according to the specific requirements of the downstream task, balancing the benefits of pretrained representations with task-specific adaptability.

\subsection{Limitation and future work}

\noindent\textbf{Instance-aware features.}
In this work, we observe that pretraining is particularly beneficial for semantic-oriented tasks. For instance-discriminative tasks, pretraining also provides gains, but primarily in low-data regimes and often requires end-to-end fine-tuning. When large-scale labelled data is available, the benefits of pretrained representations diminish and may even hinder effective adaptation.

This observation is consistent with our PCA visualizations of the learned representations through self-supervised pretraining in \Cref{fig:pca_features}, which suggest that the model predominantly captures global, shared semantic features rather than instance-discriminative cues. We attribute this behavior to the current pretraining objective, which biases the model toward learning high-level shared semantics without sufficiently preserving fine-grained, instance-level distinctions~\citep{yang2026towards}. In the future work, we plan to design improved pretraining strategies that explicitly encourage instance-aware representations, thereby better supporting a wider range of downstream tasks.

\noindent\textbf{Robustness to arbitrary grid sizes.} 
We use LitePT as the backbone for pretraining. 
Following common practice in modern point cloud processing, the input point cloud is first discretized via grid sampling to obtain a fixed spatial resolution before being processed by the neural network. This step reduces the number of points and standardizes point density.

However, this design introduces an implicit constraint during downstream adaptation: using the same grid size as in pretraining typically yields the best performance, whereas deviations in grid size (and thus point density) can reduce the benefits of pretraining. 
In the future work, we plan to mitigate this by incorporating strategies such as random grid-size augmentation during pretraining, with the goal of improving robustness to varying point densities.

\noindent\textbf{Geographic generalization.} Our downstream evaluations are primarily conducted on datasets from Europe, reflecting the geographic bias of currently available and well-established benchmarks for forest point cloud understanding. Moreover, the majority of our pretraining data also originates from Europe. Consequently, the generalization of the learned representations to forests in other geographic regions remains insufficiently evaluated. Ideally, the model should be assessed on datasets from diverse regions, such as Asia, Africa, or South America. However, such evaluation is currently limited by the lack of well-established forest point cloud benchmarks with suitable annotations in these regions. Expanding both pretraining data and downstream evaluations to more geographically diverse forests is therefore an important direction for future work.

\section{Conclusion}
\label{sec_conclusion}

This work presents a systematic study of large-scale representation learning for forest point clouds through both self-supervised and supervised pretraining. We show that self-supervised pretraining accelerates model convergence and improves downstream performance under limited annotation budgets. While task-specific supervised pretraining benefits closely related downstream tasks, self-supervised pretraining learns more transferable representations, providing the strongest foundation for cross-task transfer. Together, these findings suggest that large-scale self-supervised pretraining is a promising paradigm for forest point cloud understanding. At the same time, our results indicate that learning strongly instance-discriminative representations remains an important open challenge for current pretraining methods. Future work will explore scaling pretraining to global-scale and more diverse forest datasets, enhancing representation learning for instance-level discrimination, improving robustness to varying point densities and acquisition conditions, and extending pretrained representations to a broader range of ecological applications. Ultimately, we envision foundation models that provide a unified representation for general-purpose 3D forest understanding.

\section*{Acknowledgments} 
Part of the compute is supported by the Swiss AI Initiative under projects a0182 and a144 on the Alps supercomputer. 

The project is supported by the Circular Bio-based Europe Joint Undertaking (CBE JU) and its members under Grant Agreement No 101157488 (SingleTree). This project is also funded by the EU Horizon Europe program under Grant Agreement No 101213369 (DVPS) and Grant Agreement No 101131841 (Embed2Scale). Additional funding for Embed2Scale has been provided by the Swiss State Secretariat for Education, Research and Innovation (SERI) and UK Research and Innovation (UKRI).

Funded by the European Union. Views and opinions expressed are however those of the author(s) only and do not necessarily reflect those of the European Union or CBE JU. Neither the European Union nor the CBE JU can be held responsible for them.

\bibliographystyle{abbrvnat}
\bibliography{references}

\begin{thebibliography}{75}
\providecommand{\natexlab}[1]{#1}
\providecommand{\url}[1]{\texttt{#1}}
\expandafter\ifx\csname urlstyle\endcsname\relax
  \providecommand{\doi}[1]{doi: #1}\else
  \providecommand{\doi}{doi: \begingroup \urlstyle{rm}\Url}\fi

\bibitem[Achiam et~al.(2023)Achiam, Adler, Agarwal, Ahmad, Akkaya, Aleman, Almeida, Altenschmidt, Altman, Anadkat, et~al.]{achiam2023gpt}
J.~Achiam, S.~Adler, S.~Agarwal, L.~Ahmad, I.~Akkaya, F.~L. Aleman, D.~Almeida, J.~Altenschmidt, S.~Altman, S.~Anadkat, et~al.
\newblock {GPT-4} technical report.
\newblock arXiv preprint, 2023.

\bibitem[Anderegg et~al.(2022)Anderegg, Wu, Acil, Carvalhais, Pugh, Sadler, and Seidl]{anderegg2022climate}
W.~R. Anderegg, C.~Wu, N.~Acil, N.~Carvalhais, T.~A. Pugh, J.~P. Sadler, and R.~Seidl.
\newblock A climate risk analysis of {Earth’s} forests in the 21st century.
\newblock \emph{Science}, 377\penalty0 (6610):\penalty0 1099--1103, 2022.

\bibitem[Astruc et~al.(2025)Astruc, Gonthier, Mallet, and Landrieu]{astruc2025anysat}
G.~Astruc, N.~Gonthier, C.~Mallet, and L.~Landrieu.
\newblock {Anysat}: One earth observation model for many resolutions, scales, and modalities.
\newblock In \emph{IEEE/CVF Conference on Computer Vision and Pattern Recognition (CVPR)}, 2025.

\bibitem[Ba et~al.(2016)Ba, Kiros, and Hinton]{ba2016layer}
J.~L. Ba, J.~R. Kiros, and G.~E. Hinton.
\newblock {Layer Normalization}.
\newblock \emph{arXiv preprint arXiv:1607.06450}, 2016.

\bibitem[Berman et~al.(2018)Berman, Triki, and Blaschko]{berman2018lovasz}
M.~Berman, A.~R. Triki, and M.~B. Blaschko.
\newblock The {Lovász-Softmax} loss: A tractable surrogate for the optimization of the intersection-over-union measure in neural networks.
\newblock In \emph{IEEE/CVF Conference on Computer Vision and Pattern Recognition (CVPR)}, 2018.

\bibitem[Bodnar et~al.(2025)Bodnar, Bruinsma, Lucic, Stanley, Allen, Brandstetter, Garvan, Riechert, Weyn, Dong, et~al.]{bodnar2025foundation}
C.~Bodnar, W.~P. Bruinsma, A.~Lucic, M.~Stanley, A.~Allen, J.~Brandstetter, P.~Garvan, M.~Riechert, J.~A. Weyn, H.~Dong, et~al.
\newblock A foundation model for the {Earth} system.
\newblock \emph{Nature}, 641:\penalty0 1180--1187, 2025.

\bibitem[Borsah et~al.(2023)Borsah, Nazeer, and Wong]{borsah2023lidar}
A.~A. Borsah, M.~Nazeer, and M.~S. Wong.
\newblock {LiDAR}-based forest biomass remote sensing: A review of metrics, methods, and assessment criteria for the selection of allometric equations.
\newblock \emph{Forests}, 14\penalty0 (10):\penalty0 2095, 2023.

\bibitem[Bountos et~al.(2025)Bountos, Ouaknine, Papoutsis, and Rolnick]{bountos2025fomo}
N.~I. Bountos, A.~Ouaknine, I.~Papoutsis, and D.~Rolnick.
\newblock Fomo: Multi-modal, multi-scale and multi-task remote sensing foundation models for forest monitoring.
\newblock In \emph{Conference on Artificial Intelligence (AAAI)}, 2025.

\bibitem[Brown et~al.(2025)Brown, Kazmierski, Pasquarella, Rucklidge, Samsikova, Zhang, Shelhamer, Lahera, Wiles, Ilyushchenko, et~al.]{brown2025alphaearth}
C.~F. Brown, M.~R. Kazmierski, V.~J. Pasquarella, W.~J. Rucklidge, M.~Samsikova, C.~Zhang, E.~Shelhamer, E.~Lahera, O.~Wiles, S.~Ilyushchenko, et~al.
\newblock Alphaearth foundations: An embedding field model for accurate and efficient global mapping from sparse label data.
\newblock arXiv preprint, 2025.

\bibitem[Cherlet et~al.(2026)Cherlet, Dayal, Chen, Cooper, Disney, Hanzl, Levick, Nightingale, Origo, Senf, et~al.]{cherlet2026benchmarking}
W.~Cherlet, K.~Dayal, S.~Chen, Z.~Cooper, M.~Disney, A.~Hanzl, S.~Levick, J.~Nightingale, N.~Origo, C.~Senf, et~al.
\newblock Benchmarking tree instance segmentation of terrestrial laser scanning point clouds.
\newblock \emph{ISPRS Journal of Photogrammetry and Remote Sensing}, 231:\penalty0 230--247, 2026.

\bibitem[Choy et~al.(2019)Choy, Gwak, and Savarese]{choy20194d}
C.~Choy, J.~Gwak, and S.~Savarese.
\newblock {4D} spatio-temporal {ConvNets}: Minkowski convolutional neural networks.
\newblock In \emph{IEEE/CVF Conference on Computer Vision and Pattern Recognition (CVPR)}, 2019.

\bibitem[Comanici et~al.(2025)Comanici, Bieber, Schaekermann, Pasupat, Sachdeva, Dhillon, Blistein, Ram, Zhang, Rosen, et~al.]{comanici2025gemini}
G.~Comanici, E.~Bieber, M.~Schaekermann, I.~Pasupat, N.~Sachdeva, I.~Dhillon, M.~Blistein, O.~Ram, D.~Zhang, E.~Rosen, et~al.
\newblock Gemini 2.5: Pushing the frontier with advanced reasoning, multimodality, long context, and next generation agentic capabilities.
\newblock arXiv preprint, 2025.

\bibitem[Duguay et~al.(2026)Duguay, Baudchon, Lalibert{\'e}, Muller-Landau, Rivas-Torres, and Ouaknine]{duguay2026selvamask}
S.-O. Duguay, H.~Baudchon, E.~Lalibert{\'e}, H.~Muller-Landau, G.~Rivas-Torres, and A.~Ouaknine.
\newblock {SelvaMask}: Segmenting trees in tropical forests and beyond.
\newblock arXiv preprint, 2026.

\bibitem[Duncanson et~al.(2022)Duncanson, Kellner, Armston, Dubayah, Minor, Hancock, Healey, Patterson, Saarela, Marselis, Silva, Bruening, Goetz, Tang, Hofton, Blair, Luthcke, Fatoyinbo, et~al.]{duncanson2022_gedi_agbd}
L.~Duncanson, J.~R. Kellner, J.~Armston, R.~Dubayah, D.~M. Minor, S.~Hancock, S.~P. Healey, P.~L. Patterson, S.~Saarela, S.~Marselis, C.~E. Silva, J.~Bruening, S.~J. Goetz, H.~Tang, M.~Hofton, B.~Blair, S.~Luthcke, L.~Fatoyinbo, et~al.
\newblock Aboveground biomass density models for {NASA}’s {Global} {Ecosystem} {Dynamics} {Investigation} ({GEDI}) {LiDAR} mission.
\newblock \emph{Remote Sensing of Environment}, 270:\penalty0 112845, 2022.

\bibitem[Fareed et~al.(2026)Fareed, Silva, Izaya, and Flores]{nadeem2026interdisciplinary}
N.~Fareed, C.~A. Silva, N.~Izaya, and J.~P. Flores.
\newblock Interdisciplinary applications of {LiDAR} in forest studies: Advances in sensors, methods, and cross-domain metrics.
\newblock \emph{Remote Sensing}, 18\penalty0 (2):\penalty0 219, 2026.

\bibitem[Fogel et~al.(2025)Fogel, Perron, Besic, Saint-Andr{\'e}, Pellissier-Tanon, Schwartz, Boudras, Fayad, d'Aspremont, Landrieu, et~al.]{fogel2025open}
F.~Fogel, Y.~Perron, N.~Besic, L.~Saint-Andr{\'e}, A.~Pellissier-Tanon, M.~Schwartz, T.~Boudras, I.~Fayad, A.~d'Aspremont, L.~Landrieu, et~al.
\newblock Open-canopy: Towards very high resolution forest monitoring.
\newblock In \emph{IEEE/CVF Conference on Computer Vision and Pattern Recognition (CVPR)}, 2025.

\bibitem[Forzieri et~al.(2022)Forzieri, Dakos, McDowell, Ramdane, and Cescatti]{forzieri2022emerging}
G.~Forzieri, V.~Dakos, N.~G. McDowell, A.~Ramdane, and A.~Cescatti.
\newblock Emerging signals of declining forest resilience under climate change.
\newblock \emph{Nature}, 608\penalty0 (7923):\penalty0 534--539, 2022.

\bibitem[Gaydon and Roche(2025)]{gaydon2025pureforest}
C.~Gaydon and F.~Roche.
\newblock Pureforest: A large-scale aerial {LiDAR} and aerial imagery dataset for tree species classification in monospecific forests.
\newblock In \emph{2025 IEEE/CVF Winter Conference on Applications of Computer Vision (WACV)}, 2025.

\bibitem[Geist et~al.(2026)Geist, Landrieu, and Robert]{geist2026ezsp}
L.~Geist, L.~Landrieu, and D.~Robert.
\newblock {EZ-SP}: Fast and lightweight superpoint-based {3D} segmentation.
\newblock In \emph{International Conference on Robotics and Automation (ICRA)}, 2026.

\bibitem[Graham et~al.(2018)Graham, Engelcke, and Van Der~Maaten]{graham20183d}
B.~Graham, M.~Engelcke, and L.~Van Der~Maaten.
\newblock {3D} semantic segmentation with submanifold sparse convolutional networks.
\newblock In \emph{IEEE/CVF Conference on Computer Vision and Pattern Recognition (CVPR)}, 2018.

\bibitem[Guo et~al.(2024)Guo, Lao, Dang, Zhang, Yu, Ru, Zhong, Huang, Wu, Hu, et~al.]{guo2024skysense}
X.~Guo, J.~Lao, B.~Dang, Y.~Zhang, L.~Yu, L.~Ru, L.~Zhong, Z.~Huang, K.~Wu, D.~Hu, et~al.
\newblock {Skysense}: A multi-modal remote sensing foundation model towards universal interpretation for earth observation imagery.
\newblock In \emph{IEEE/CVF Conference on Computer Vision and Pattern Recognition (CVPR)}, 2024.

\bibitem[He et~al.(2022)He, Chen, Xie, Li, Doll{\'a}r, and Girshick]{he2022masked}
K.~He, X.~Chen, S.~Xie, Y.~Li, P.~Doll{\'a}r, and R.~Girshick.
\newblock Masked autoencoders are scalable vision learners.
\newblock In \emph{IEEE/CVF Conference on Computer Vision and Pattern Recognition (CVPR)}, 2022.

\bibitem[Henrich et~al.(2024)Henrich, van Delden, Seidel, Kneib, and Ecker]{henrich2024treelearn}
J.~Henrich, J.~van Delden, D.~Seidel, T.~Kneib, and A.~S. Ecker.
\newblock {TreeLearn}: A deep learning method for segmenting individual trees from ground-based {LiDAR} forest point clouds.
\newblock \emph{Ecological Informatics}, 84:\penalty0 102888, 2024.

\bibitem[Hua et~al.(2022)Hua, Bruijnzeel, Meli, Martin, Zhang, Nakagawa, Miao, Wang, McEvoy, Pe{\~n}a-Arancibia, et~al.]{hua2022biodiversity}
F.~Hua, L.~A. Bruijnzeel, P.~Meli, P.~A. Martin, J.~Zhang, S.~Nakagawa, X.~Miao, W.~Wang, C.~McEvoy, J.~L. Pe{\~n}a-Arancibia, et~al.
\newblock The biodiversity and ecosystem service contributions and trade-offs of forest restoration approaches.
\newblock \emph{Science}, 376\penalty0 (6595):\penalty0 839--844, 2022.

\bibitem[Ioffe and Szegedy(2015)]{ioffe2015batch}
S.~Ioffe and C.~Szegedy.
\newblock {Batch Normalization}: Accelerating deep network training by reducing internal covariate shift.
\newblock In \emph{International Conference on Machine Learning (ICML)}, 2015.

\bibitem[Jakubik et~al.(2025)Jakubik, Yang, Blumenstiel, Scheurer, Sedona, Maurogiovanni, Bosmans, Dionelis, Marsocci, Kopp, et~al.]{jakubik2025terramind}
J.~Jakubik, F.~Yang, B.~Blumenstiel, E.~Scheurer, R.~Sedona, S.~Maurogiovanni, J.~Bosmans, N.~Dionelis, V.~Marsocci, N.~Kopp, et~al.
\newblock {TerraMind}: Large-scale generative multimodality for earth observation.
\newblock In \emph{International Conference on Computer Vision (ICCV)}, 2025.

\bibitem[Jiang et~al.(2026)Jiang, Shen, Wang, Kissling, Hollaus, Su, Wang, Ferreira, and Pfeifer]{jiang2026cross}
J.~Jiang, Y.~Shen, J.~Wang, W.~D. Kissling, M.~Hollaus, H.~Su, J.~Wang, V.~Ferreira, and N.~Pfeifer.
\newblock Cross-platform forest understanding: A multi-platform synergistic training framework for generalized forest point cloud segmentation.
\newblock \emph{Remote Sensing of Environment}, 342:\penalty0 115467, 2026.

\bibitem[Kirillov et~al.(2023)Kirillov, Mintun, Ravi, Mao, Rolland, Gustafson, Xiao, Whitehead, Berg, Lo, et~al.]{kirillov2023segment}
A.~Kirillov, E.~Mintun, N.~Ravi, H.~Mao, C.~Rolland, L.~Gustafson, T.~Xiao, S.~Whitehead, A.~C. Berg, W.-Y. Lo, et~al.
\newblock Segment anything.
\newblock In \emph{International Conference on Computer Vision (ICCV)}, 2023.

\bibitem[Kolodiazhnyi et~al.(2024)Kolodiazhnyi, Vorontsova, Konushin, and Rukhovich]{kolodiazhnyi2024oneformer3d}
M.~Kolodiazhnyi, A.~Vorontsova, A.~Konushin, and D.~Rukhovich.
\newblock {OneFormer3D}: One transformer for unified point cloud segmentation.
\newblock In \emph{IEEE/CVF Conference on Computer Vision and Pattern Recognition (CVPR)}, 2024.

\bibitem[Laino et~al.(2026)Laino, Cabo, Ord{\'o}{\~n}ez, Bolanos, Janvier, Giulioni, Herrmann, Hudak, Parsons, and Santin]{laino2025_segmentedforests}
D.~Laino, C.~Cabo, C.~Ord{\'o}{\~n}ez, R.~Bolanos, R.~Janvier, F.~Giulioni, M.~Herrmann, A.~Hudak, R.~Parsons, and C.~Santin.
\newblock {SegmentedForests}: A labelled dataset of terrestrial {LiDAR} point clouds for semantic segmentation of forests.
\newblock \emph{Forestry: An International Journal of Forest Research}, 99\penalty0 (2):\penalty0 cpaf062, 2026.

\bibitem[Lang et~al.(2023)Lang, Jetz, Schindler, and Wegner]{lang2023high}
N.~Lang, W.~Jetz, K.~Schindler, and J.~D. Wegner.
\newblock A high-resolution canopy height model of the earth.
\newblock \emph{Nature Ecology \& Evolution}, 7\penalty0 (11):\penalty0 1778--1789, 2023.

\bibitem[Lin et~al.(2014)Lin, Maire, Belongie, Hays, Perona, Ramanan, Doll{\'a}r, and Zitnick]{lin2014microsoft}
T.-Y. Lin, M.~Maire, S.~Belongie, J.~Hays, P.~Perona, D.~Ramanan, P.~Doll{\'a}r, and C.~L. Zitnick.
\newblock Microsoft {COCO}: Common objects in context.
\newblock In \emph{European Conference on Computer Vision (ECCV)}, 2014.

\bibitem[Liu et~al.(2024)Liu, Feng, Xue, Wang, Wu, Lu, Zhao, Deng, Zhang, Ruan, et~al.]{liu2024deepseek}
A.~Liu, B.~Feng, B.~Xue, B.~Wang, B.~Wu, C.~Lu, C.~Zhao, C.~Deng, C.~Zhang, C.~Ruan, et~al.
\newblock Deepseek-v3 technical report.
\newblock arXiv preprint, 2024.

\bibitem[Liu et~al.(2026)Liu, Wang, Gong, Wang, Zhu, and Wang]{liu2026synthetic}
J.~Liu, D.~Wang, H.~Gong, C.~Wang, J.~Zhu, and D.~Wang.
\newblock A synthetic data generation framework for deep learning-based {LiDAR} forest structure analysis.
\newblock \emph{Remote Sensing of Environment}, 341:\penalty0 115436, 2026.

\bibitem[Liu et~al.(2022)Liu, Ma, Wu, Hu, Liu, Liu, Guo, and Su]{liu2022novel}
X.~Liu, Q.~Ma, X.~Wu, T.~Hu, Z.~Liu, L.~Liu, Q.~Guo, and Y.~Su.
\newblock A novel entropy-based method to quantify forest canopy structural complexity from multiplatform {LiDAR} point clouds.
\newblock \emph{Remote Sensing of Environment}, 282:\penalty0 113280, 2022.

\bibitem[Loshchilov and Hutter(2019)]{loshchilov2017decoupled}
I.~Loshchilov and F.~Hutter.
\newblock Decoupled weight decay regularization.
\newblock \emph{International Conference on Learning Representations (ICLR)}, 2019.

\bibitem[Lu et~al.(2025)Lu, Li, Yang, Fan, Wang, Pang, Wang, Lian, Xu, and Huang]{lu2025towards}
H.~Lu, B.~Li, G.~Yang, G.~Fan, H.~Wang, Y.~Pang, Z.~Wang, Y.~Lian, H.~Xu, and H.~Huang.
\newblock Towards a point cloud understanding framework for forest scene semantic segmentation across forest types and sensor platforms.
\newblock \emph{Remote Sensing of Environment}, 2025.

\bibitem[Maeda et~al.(2025)Maeda, Brede, Calders, Disney, Herold, Lines, Nunes, Raumonen, Rautiainen, Saarinen, et~al.]{maeda2025expanding}
E.~E. Maeda, B.~Brede, K.~Calders, M.~Disney, M.~Herold, E.~R. Lines, M.~H. Nunes, P.~Raumonen, M.~Rautiainen, N.~Saarinen, et~al.
\newblock Expanding forest research with terrestrial {LiDAR} technology.
\newblock \emph{Nature Communications}, 16\penalty0 (1):\penalty0 8853, 2025.

\bibitem[Micha{\l}owska and Rapi{\'n}ski(2021)]{michalowska2021review}
M.~Micha{\l}owska and J.~Rapi{\'n}ski.
\newblock A review of tree species classification based on airborne {LiDAR} data and applied classifiers.
\newblock \emph{Remote Sensing}, 13\penalty0 (3):\penalty0 353, 2021.

\bibitem[Migliavacca et~al.(2021)Migliavacca, Musavi, Mahecha, Nelson, Knauer, Baldocchi, Perez-Priego, Christiansen, Peters, Anderson, et~al.]{migliavacca2021three}
M.~Migliavacca, T.~Musavi, M.~D. Mahecha, J.~A. Nelson, J.~Knauer, D.~D. Baldocchi, O.~Perez-Priego, R.~Christiansen, J.~Peters, K.~Anderson, et~al.
\newblock The three major axes of terrestrial ecosystem function.
\newblock \emph{Nature}, 598:\penalty0 468--472, 2021.

\bibitem[Mo et~al.(2023)Mo, Zohner, Reich, Liang, De~Miguel, Nabuurs, Renner, Van Den~Hoogen, Araza, Herold, et~al.]{mo2023integrated}
L.~Mo, C.~M. Zohner, P.~B. Reich, J.~Liang, S.~De~Miguel, G.-J. Nabuurs, S.~S. Renner, J.~Van Den~Hoogen, A.~Araza, M.~Herold, et~al.
\newblock Integrated global assessment of the natural forest carbon potential.
\newblock \emph{Nature}, 624\penalty0 (7990):\penalty0 92--101, 2023.

\bibitem[Nguyen et~al.(2026)Nguyen, Vu, Le, Kawanishi, Komamizu, Ide, and Kattenborn]{nguyen2026forestmamba}
T.~T. Nguyen, T.-A. Vu, D.~V. Le, Y.~Kawanishi, T.~Komamizu, I.~Ide, and T.~Kattenborn.
\newblock {ForestMamba}: Sparse mamba with geometry-guided queries for {3D} forest point cloud segmentation.
\newblock In \emph{British Machine Vision Conference (BMVC)}, 2026.

\bibitem[Oehmcke et~al.(2024)Oehmcke, Li, Trepekli, Revenga, Nord-Larsen, Gieseke, and Igel]{oehmcke2024deep}
S.~Oehmcke, L.~Li, K.~Trepekli, J.~C. Revenga, T.~Nord-Larsen, F.~Gieseke, and C.~Igel.
\newblock Deep point cloud regression for above-ground forest biomass estimation from airborne {LiDAR}.
\newblock \emph{Remote Sensing of Environment}, 302:\penalty0 113968, 2024.

\bibitem[Opler et~al.(2026)Opler, Ciais, Fayad, Schwartz, Belouze, Brood, D’Aspremont, Gominski, Aubry, and Landrieu]{opler2026prototree}
A.~Opler, P.~Ciais, I.~Fayad, M.~Schwartz, G.~Belouze, S.~Brood, A.~D’Aspremont, D.~Gominski, M.~Aubry, and L.~Landrieu.
\newblock {ProtoTree}: An efficient and generalizable model for individual tree point clouds analysis.
\newblock \emph{IEEE Transactions on Geoscience and Remote Sensing}, 2026.

\bibitem[Oquab et~al.(2024)Oquab, Darcet, Moutakanni, Vo, Szafraniec, Khalidov, Fernandez, Haziza, Massa, El-Nouby, et~al.]{oquab2023dinov2}
M.~Oquab, T.~Darcet, T.~Moutakanni, H.~Vo, M.~Szafraniec, V.~Khalidov, P.~Fernandez, D.~Haziza, F.~Massa, A.~El-Nouby, et~al.
\newblock {DINOv2}: Learning robust visual features without supervision.
\newblock \emph{Transactions on Machine Learning Research}, 2024.

\bibitem[Pan et~al.(2024)Pan, Birdsey, Phillips, Houghton, Fang, Kauppi, Keith, Kurz, Ito, Lewis, Nabuurs, Shvidenko, Hashimoto, Lerink, Schepaschenko, Castanho, and Murdiyarso]{pan2024enduring}
Y.~Pan, R.~A. Birdsey, O.~L. Phillips, R.~A. Houghton, J.~Fang, P.~E. Kauppi, H.~Keith, W.~A. Kurz, A.~Ito, S.~L. Lewis, G.-J. Nabuurs, A.~Shvidenko, S.~Hashimoto, B.~Lerink, D.~Schepaschenko, A.~Castanho, and D.~Murdiyarso.
\newblock The enduring world forest carbon sink.
\newblock \emph{Nature}, 631\penalty0 (8021):\penalty0 563--569, 2024.

\bibitem[Plekhanova et~al.(2025)Plekhanova, Robert, Dollinger, Arens, Brun, Wegner, and Zimmermann]{plekhanova2025ssl4eco}
E.~Plekhanova, D.~Robert, J.~Dollinger, E.~Arens, P.~Brun, J.~D. Wegner, and N.~E. Zimmermann.
\newblock {SSL4Eco}: A global seasonal dataset for geospatial foundation models in ecology.
\newblock In \emph{CVPR EarthVision Workshop}, 2025.

\bibitem[Puliti et~al.(2023{\natexlab{a}})Puliti, Pearse, Surov{\'y}, Wallace, Hollaus, Wielgosz, and Astrup]{puliti2023_forinstance_data}
S.~Puliti, G.~Pearse, P.~Surov{\'y}, L.~Wallace, M.~Hollaus, M.~Wielgosz, and R.~Astrup.
\newblock {FOR-instance}: A {UAV} laser scanning benchmark dataset for semantic and instance segmentation of individual trees (dataset).
\newblock Zenodo, 2023{\natexlab{a}}.

\bibitem[Puliti et~al.(2023{\natexlab{b}})Puliti, Pearse, Surov{\'y}, Wallace, Hollaus, Wielgosz, and Astrup]{puliti2023_forinstance_paper}
S.~Puliti, G.~Pearse, P.~Surov{\'y}, L.~Wallace, M.~Hollaus, M.~Wielgosz, and R.~Astrup.
\newblock {FOR-instance}: A {UAV} laser scanning benchmark dataset for semantic and instance segmentation of individual trees.
\newblock arXiv preprint, 2023{\natexlab{b}}.

\bibitem[Puliti et~al.(2025)Puliti, Lines, M{\"u}llerov{\'a}, Frey, Schindler, Straker, Allen, Winiwarter, Rehush, Hristova, Murray, Calders, Terryn, Coops, H{\"o}fle, Junttila, Kr{\r{u}}{\v{c}}ek, Krok, Kr{\'a}l, Levick, Luck, Missarov, Mokro{\v{s}}, Owen, Stere{\'n}czak, Pitk{\"a}nen, Puletti, Saarinen, Hopkinson, Torresan, Tomelleri, Weiser, and Astrup]{puliti2025_forspecies20k}
S.~Puliti, E.~R. Lines, J.~M{\"u}llerov{\'a}, J.~Frey, Z.~Schindler, A.~Straker, M.~J. Allen, L.~Winiwarter, N.~Rehush, H.~Hristova, B.~Murray, K.~Calders, L.~Terryn, N.~Coops, B.~H{\"o}fle, S.~Junttila, M.~Kr{\r{u}}{\v{c}}ek, G.~Krok, K.~Kr{\'a}l, S.~R. Levick, L.~Luck, A.~Missarov, M.~Mokro{\v{s}}, H.~J.~F. Owen, K.~Stere{\'n}czak, T.~P. Pitk{\"a}nen, N.~Puletti, N.~Saarinen, C.~Hopkinson, C.~Torresan, E.~Tomelleri, H.~Weiser, and R.~Astrup.
\newblock Benchmarking tree species classification from proximally sensed laser scanning data: Introducing the {FOR-species20K} dataset.
\newblock \emph{Methods in Ecology and Evolution}, 16:\penalty0 801--818, 2025.

\bibitem[Puliti et~al.(2026)Puliti, Xiang, Wielgosz, Handegard, Cattaneo, Vergarechea, Gobakken, Hyyppä, Næsset, Vastaranta, Yrttimaa, and Astrup]{PULITI2026115462}
S.~Puliti, B.~Xiang, M.~Wielgosz, E.~Handegard, N.~Cattaneo, M.~Vergarechea, T.~Gobakken, J.~Hyyppä, E.~Næsset, M.~Vastaranta, T.~Yrttimaa, and R.~Astrup.
\newblock {FOR-age}: Benchmarking individual tree age estimation using {3D} deep learning on dense laser scanning data.
\newblock \emph{Remote Sensing of Environment}, 342:\penalty0 115462, 2026.
\newblock ISSN 0034-4257.

\bibitem[Radford et~al.(2021)Radford, Kim, Hallacy, Ramesh, Goh, Agarwal, Sastry, Askell, Mishkin, Clark, et~al.]{radford2021learning}
A.~Radford, J.~W. Kim, C.~Hallacy, A.~Ramesh, G.~Goh, S.~Agarwal, G.~Sastry, A.~Askell, P.~Mishkin, J.~Clark, et~al.
\newblock Learning transferable visual models from natural language supervision.
\newblock In \emph{International Conference on Machine Learning (ICML)}, 2021.

\bibitem[Rizaldy et~al.(2026)Rizaldy, Fassnacht, Afifi, Jiang, Gloaguen, and Ghamisi]{rizaldy2026label}
A.~Rizaldy, F.~E. Fassnacht, A.~J. Afifi, H.~Jiang, R.~Gloaguen, and P.~Ghamisi.
\newblock Label-efficient {3D} forest mapping: Self-supervised and transfer learning for instance segmentation, semantic segmentation, and species classification.
\newblock \emph{Remote Sensing of Environment}, 345:\penalty0 115564, 2026.
\newblock ISSN 0034-4257.

\bibitem[Robert et~al.(2023)Robert, Raguet, and Landrieu]{robert2023efficient}
D.~Robert, H.~Raguet, and L.~Landrieu.
\newblock Efficient 3d semantic segmentation with superpoint transformer.
\newblock In \emph{International Conference on Computer Vision (ICCV)}, 2023.

\bibitem[Shao et~al.(2024)Shao, Lin, Wingren, Shin, Fei, Carpenter, Habib, and Fei]{shao2024large}
J.~Shao, Y.-C. Lin, C.~Wingren, S.-Y. Shin, W.~Fei, J.~Carpenter, A.~Habib, and S.~Fei.
\newblock Large-scale inventory in natural forests with mobile {LiDAR} point clouds.
\newblock \emph{Science of Remote Sensing}, 2024.

\bibitem[She et~al.(2026)She, Blake, Coomes, and Keshav]{she2026scaling}
Y.~She, A.~Blake, D.~Coomes, and S.~Keshav.
\newblock Scaling up forest vision with synthetic data.
\newblock \emph{International Journal on Computer Vision (IJCV)}, 134\penalty0 (7):\penalty0 343, 2026.

\bibitem[Sim{\'e}oni et~al.(2025)Sim{\'e}oni, Vo, Seitzer, Baldassarre, Oquab, Jose, Khalidov, Szafraniec, Yi, Ramamonjisoa, et~al.]{simeoni2025dinov3}
O.~Sim{\'e}oni, H.~V. Vo, M.~Seitzer, F.~Baldassarre, M.~Oquab, C.~Jose, V.~Khalidov, M.~Szafraniec, S.~Yi, M.~Ramamonjisoa, et~al.
\newblock {DINOv3}.
\newblock arXiv preprint, 2025.

\bibitem[Skidmore et~al.(2021)Skidmore, Coops, Neinavaz, Ali, Schaepman, Paganini, Kissling, Vihervaara, Darvishzadeh, Feilhauer, et~al.]{skidmore2021priority}
A.~K. Skidmore, N.~C. Coops, E.~Neinavaz, A.~Ali, M.~E. Schaepman, M.~Paganini, W.~D. Kissling, P.~Vihervaara, R.~Darvishzadeh, H.~Feilhauer, et~al.
\newblock Priority list of biodiversity metrics to observe from space.
\newblock \emph{Nature Ecology \& Evolution}, 5\penalty0 (7):\penalty0 896--906, 2021.

\bibitem[Smith and Topin(2019)]{smith2019super}
L.~N. Smith and N.~Topin.
\newblock Super-convergence: Very fast training of neural networks using large learning rates.
\newblock In \emph{Artificial Intelligence and Machine Learning for Multi-Domain Operations Applications}, 2019.

\bibitem[Szwarcman et~al.(2025)Szwarcman, Roy, Fraccaro, G{\'\i}slason, Blumenstiel, Ghosal, De~Oliveira, de~Sousa~Almeida, Sedona, Kang, et~al.]{szwarcman2025prithvi}
D.~Szwarcman, S.~Roy, P.~Fraccaro, O.~E. G{\'\i}slason, B.~Blumenstiel, R.~Ghosal, P.~H. De~Oliveira, J.~L. de~Sousa~Almeida, R.~Sedona, Y.~Kang, et~al.
\newblock Prithvi-eo-2.0: A versatile multi-temporal foundation model for earth observation applications.
\newblock \emph{IEEE Transactions on Geoscience and Remote Sensing}, 64:\penalty0 4400120, 2025.

\bibitem[Teng et~al.(2025)Teng, Ouaknine, Lalibert{\'e}, Bengio, Rolnick, and Larochelle]{teng2025sam}
M.~Teng, A.~Ouaknine, E.~Lalibert{\'e}, Y.~Bengio, D.~Rolnick, and H.~Larochelle.
\newblock Bringing {SAM} to new heights: Leveraging elevation data for tree crown segmentation from drone imagery.
\newblock arXiv preprint, 2025.

\bibitem[Tian et~al.(2025)Tian, Zhao, Meng, Sun, Zhang, Shen, Wang, Liu, and Li]{tian2025vision}
Y.~Tian, F.~Zhao, R.~Meng, R.~Sun, Y.~Zhang, Y.~Shen, B.~Wang, J.~Liu, and M.~Li.
\newblock A vision foundation model-based method for large-scale forest disturbance mapping using time series {Sentinel-1} {SAR} data.
\newblock \emph{Remote Sensing of Environment}, 325:\penalty0 114775, 2025.

\bibitem[Tolan et~al.(2024)Tolan, Yang, Nosarzewski, Couairon, Vo, Brandt, Spore, Majumdar, Haziza, Vamaraju, et~al.]{tolan2024very}
J.~Tolan, H.-I. Yang, B.~Nosarzewski, G.~Couairon, H.~V. Vo, J.~Brandt, J.~Spore, S.~Majumdar, D.~Haziza, J.~Vamaraju, et~al.
\newblock Very high resolution canopy height maps from {RGB} imagery using self-supervised vision transformer and convolutional decoder trained on aerial {LiDAR}.
\newblock \emph{Remote Sensing of Environment}, 300:\penalty0 113888, 2024.

\bibitem[Wielgosz et~al.(2024)Wielgosz, Puliti, Xiang, Schindler, and Astrup]{wielgosz2024_segmentanytree}
M.~Wielgosz, S.~Puliti, B.~Xiang, K.~Schindler, and R.~Astrup.
\newblock {SegmentAnyTree}: A sensor and platform agnostic deep learning model for tree segmentation using laser scanning data.
\newblock \emph{Remote Sensing of Environment}, 313:\penalty0 114367, 2024.

\bibitem[Wielgosz et~al.(2026)Wielgosz, Puliti, and Astrup]{wielgosz2026segmentanytreev2}
M.~Wielgosz, S.~Puliti, and R.~Astrup.
\newblock {SegmentAnyTreeV2}: Scaling transformer-based tree instance segmentation across sensors, platforms, and forests.
\newblock \emph{arXiv preprint arXiv:2606.08206}, 2026.

\bibitem[Wu et~al.(2025)Wu, DeTone, Frost, Shen, Xie, Yang, Engel, Newcombe, Zhao, and Straub]{wu2025sonata}
X.~Wu, D.~DeTone, D.~Frost, T.~Shen, C.~Xie, N.~Yang, J.~Engel, R.~Newcombe, H.~Zhao, and J.~Straub.
\newblock Sonata: Self-supervised learning of reliable point representations.
\newblock In \emph{IEEE/CVF Conference on Computer Vision and Pattern Recognition (CVPR)}, 2025.

\bibitem[Xiang et~al.(2024)Xiang, Wielgosz, Kontogianni, Peters, Puliti, Astrup, and Schindler]{xiang2024_forainet}
B.~Xiang, M.~Wielgosz, T.~Kontogianni, T.~Peters, S.~Puliti, R.~Astrup, and K.~Schindler.
\newblock Automated forest inventory: Analysis of high-density airborne {LiDAR} point clouds with {3D} deep learning.
\newblock \emph{Remote Sensing of Environment}, 305:\penalty0 114078, 2024.

\bibitem[Xiang et~al.(2025)Xiang, Wielgosz, Puliti, Kr{\'a}l, Kr{\r{u}}{\v{c}}ek, Missarov, and Astrup]{xiang2025forestformer3d}
B.~Xiang, M.~Wielgosz, S.~Puliti, K.~Kr{\'a}l, M.~Kr{\r{u}}{\v{c}}ek, A.~Missarov, and R.~Astrup.
\newblock {ForestFormer3D}: A unified framework for end-to-end segmentation of forest {LiDAR} {3D} point clouds.
\newblock In \emph{International Conference on Computer Vision (ICCV)}, 2025.

\bibitem[Yang et~al.(2026)Yang, Abdelsamad, Zhang, and Condurache]{yang2026towards}
B.~Yang, M.~Abdelsamad, M.~Zhang, and A.~P. Condurache.
\newblock {Towards foundation models for 3D scene understanding: Instance-aware self-supervised learning for point clouds}.
\newblock In \emph{IEEE/CVF Conference on Computer Vision and Pattern Recognition (CVPR)}, 2026.

\bibitem[Yang et~al.(2024)Yang, Kang, Huang, Xu, Feng, and Zhao]{yang2024depth}
L.~Yang, B.~Kang, Z.~Huang, X.~Xu, J.~Feng, and H.~Zhao.
\newblock Depth anything: Unleashing the power of large-scale unlabeled data.
\newblock In \emph{IEEE/CVF Conference on Computer Vision and Pattern Recognition (CVPR)}, 2024.

\bibitem[Yip et~al.(2024)Yip, Liu, Wu, Hau, Lin, and Zhang]{yip2024community}
K.~H.~A. Yip, R.~Liu, J.~Wu, B.~C.~H. Hau, Y.~Lin, and H.~Zhang.
\newblock Community-based plant diversity monitoring of a dense-canopy and species-rich tropical forest using airborne {LiDAR} data.
\newblock \emph{Ecological Indicators}, 158:\penalty0 111346, 2024.

\bibitem[Yue et~al.(2026)Yue, Robert, Wang, Hong, Wegner, Rupprecht, and Schindler]{yuelitept2026}
Y.~Yue, D.~Robert, J.~Wang, S.~Hong, J.~D. Wegner, C.~Rupprecht, and K.~Schindler.
\newblock {LitePT}: Lighter yet stronger point transformer.
\newblock In \emph{IEEE/CVF Conference on Computer Vision and Pattern Recognition (CVPR)}, 2026.

\bibitem[Zhang et~al.(2024)Zhang, Fischer, Prober, Yeoh, Gosper, Zdunic, and Jucker]{zhang2024robust}
B.~Zhang, F.~J. Fischer, S.~M. Prober, P.~B. Yeoh, C.~R. Gosper, K.~Zdunic, and T.~Jucker.
\newblock Robust retrieval of forest canopy structural attributes using multi-platform airborne {LiDAR}.
\newblock \emph{Remote Sensing in Ecology and Conservation}, 10\penalty0 (6):\penalty0 725--742, 2024.

\bibitem[Zhang and Pearse(2026)]{zhang2026forest}
D.~Zhang and P.~H. Pearse.
\newblock \emph{Forest economics}.
\newblock UBC Press, 2026.

\bibitem[Zhou et~al.(2022)Zhou, Wei, Wang, Shen, Xie, Yuille, and Kong]{zhou2021ibot}
J.~Zhou, C.~Wei, H.~Wang, W.~Shen, C.~Xie, A.~Yuille, and T.~Kong.
\newblock {iBOT}: Image {BERT} pre-training with online tokenizer.
\newblock \emph{International Conference on Learning Representations (ICLR)}, 2022.

\end{thebibliography}

\clearpage
\beginsupplement

\appendix
\section{Implementation details}
\label{app_implementation}

Implementation details for downstream task training, including training configurations, optimization hyperparameters, and data augmentation strategies, are provided in~\Cref{tab:downstream_implementation_details}.
\begin{table}[ht]
\centering
\footnotesize
\setlength{\tabcolsep}{3pt}
\renewcommand{\arraystretch}{1.05}
\begin{tabular}{
  >{\raggedright\arraybackslash}m{2.2cm}
  >{\raggedright\arraybackslash}m{2.0cm}
  >{\raggedright\arraybackslash}m{6.4cm}
  >{\raggedright\arraybackslash}m{3.0cm}}
\toprule
\textbf{Task} & \textbf{Variants} & \textbf{Hyperparameters} & \textbf{Data augmentation} \\
\midrule
Forest semantic segmentation &
scratch, linear probing, decoder probing, fine-tuning &
Grid size: 5\,cm, Optimizer: AdamW~\citep{loshchilov2017decoupled}, Scheduler: OneCycleLR~\citep{smith2019super}, Learning rate: 0.002, Block LR: 0.0002, Batch size: 12, Epochs: 500, Weight decay: 0.05, Loss: CrossEntropy\,+\,Lov\'{a}sz~\citep{berman2018lovasz} &
random dropout, random rotate, random scale, random flip, random jitter, elastic distortion, cylinder crop \\
\midrule
Forest instance segmentation &
scratch, head probing, decoder probing, fine-tuning &
Grid size: 20\,cm, Optimizer: AdamW~\citep{loshchilov2017decoupled}, Scheduler: OneCycleLR~\citep{smith2019super}, Learning rate: 0.0004, Batch size: 8, Epochs: 3000, Weight decay: 0.05, Loss: ForestFormer3D~\citep{xiang2025forestformer3d} &
cylinder crop, random jitter, random flip, random rotate, random scale \\
\midrule
Tree species classification &
scratch, linear probing, fine-tuning &
Grid size: 5\,cm, Optimizer: AdamW~\citep{loshchilov2017decoupled}, Scheduler: OneCycleLR~\citep{smith2019super}, Learning rate: 0.001, Block LR: 0.0001, Batch size: 128, Epochs: 300, Weight decay: 0.01, Loss: CrossEntropy\,+\,Lov\'{a}sz~\citep{berman2018lovasz} &
random scale, random rotate, random flip, random jitter, elastic distortion \\
\midrule
Tree age regression & \multicolumn{2}{l}{%
\begin{tabular}{@{}>{\raggedright\arraybackslash}m{2.0cm}>{\raggedright\arraybackslash}m{3.0cm}>{\raggedright\arraybackslash}m{3.2cm}@{}}
scratch & Learning rate: 0.0001, Loss: weighted MSE & \multirow{3}{3.2cm}{Grid size: 5\,cm, Optimizer: AdamW~\citep{loshchilov2017decoupled}, Scheduler: PolyLR, Batch size: 128, Epochs: 9000, Weight decay: 0.05} \\
\addlinespace[1.5pt]
linear probing & Learning rate: 0.0001, Loss: Smooth L1 ($\beta=20$) & \\
\addlinespace[1.5pt]
fine-tuning & Learning rate: 0.00002, Head LR: 0.0001, Loss: weighted MSE & \\
\end{tabular}
} & random rotate, random scale, random dropout, random shift \\
\bottomrule
\end{tabular}

\caption{\textbf{Downstream task implementation details.} Training configurations, optimization hyperparameters, and data augmentation for each downstream task. All experiments were conducted using four NVIDIA GH200 GPUs to facilitate rapid experiment iteration; the use of four GPUs was not necessitated by the memory requirements of the experiments.}
\label{tab:downstream_implementation_details}
\end{table}

\end{document}